\documentclass[journal]{IEEEtran}

\usepackage{graphicx} 
\usepackage{epstopdf}
\DeclareGraphicsExtensions{.eps}

\usepackage{multirow}
\usepackage{algorithmic}
\usepackage{algorithm}
\usepackage{amsmath} 
\usepackage{amssymb}
\usepackage{amsxtra}
\usepackage{url} 
\usepackage{color} 
\usepackage{bm}
\usepackage{placeins}
\usepackage[caption=false]{subfig}
\usepackage{xspace}
\usepackage{rotating}
\usepackage{siunitx}
\usepackage{textcomp}%
\usepackage{algorithm}
\usepackage{algorithmic}

\newcommand{\added}[2]{#1}

\newcommand{\wrt}{\text{w.r.t.}\xspace}

\newcommand{\myvector}[1]{\bm{#1}}
\newcommand{\myvec}[1]{\myvector{#1}}

\newcommand{\R}[1]{\mathbb{R}^{#1}}

\newcommand{\best}[1]{\textbf{#1}}
\newcommand{\algorithmicinput}{\textbf{Input:}}
\newcommand{\algorithmicoutput}{\textbf{Output:}}
\newcommand{\INPUT}{\item[\algorithmicinput]}
\newcommand{\OUTPUT}{\item[\algorithmicoutput]}

\newcommand{\dist}[1]{\SI{#1}{\meter}}

\newcommand{\velocity}[1]{\ensuremath{#1\,{\text{m}/\text{s}}}}
\newcommand{\angvel}[1]{\ensuremath{#1\,{\text{rad}/\text{s}}}}

\newcommand{\gauss}[2]{\mathcal{N}(#1 , #2)}

\newcommand{\ceil}[1]{\lceil{#1}\rceil}

\renewcommand{\baselinestretch}{.985}

\newcommand{\excise}[1]{}

\newif\ifremark
\long\def\remark#1{
  \ifremark%
  \begingroup%
  \dimen0=\textwidth
  \advance\dimen0 by -1in%
  \setbox0=\hbox{\parbox[b]{\dimen0}{\protect\em #1}}
  \dimen1=\ht0\advance\dimen1 by 2pt%
  \dimen2=\dp0\advance\dimen2 by 2pt%
  \vskip 0.25pt%
  \hbox to \textwidth{%
    \vrule height\dimen1 width 3pt depth\dimen2%
    \hss\copy0\hss%
    \vrule height\dimen1 width 3pt depth\dimen2%
  }%
  \endgroup%
  \fi}

\newcommand{\x}{\ensuremath{\myvec{s}}}

\newcommand{\uv}{\ensuremath{\myvec{a}}}
\newcommand{\ui}[1]{a}

\newcommand{\g}{\ensuremath{\myvec{g}}}

\newcommand{\cspace}{\mathcal{C}}
\newcommand{\cfree}{\mathcal{C}_{\mathrm{free}}}
\newcommand{\tfree}{T_{\mathrm{free}}}

\newcommand{\ac}{\uv}

\ifCLASSINFOpdf
\else
\fi

\usepackage{gensymb}

\usepackage{float}

\begin{document}
\title{Long-Range Indoor Navigation with PRM-RL}

\author{Anthony~Francis \and
        Aleksandra~Faust \and
        Hao-Tien Lewis Chiang \\
        Jasmine~Hsu \and
        J.~Chase~Kew \and
        Marek~Fiser \and
         Tsang-Wei~Edward~Lee
\thanks{The authors are with Robotics at Google, 1600 Amphitheatre Parkway, Mountain View, CA, 94043 (email: centaur, lewispro, hellojas, jkew, mfiser, tsangwei, sandrafaust all @google.com).}
}

%
%

\markboth{Transactions on Robotics,~Vol.~??, No.~?, ???~2020}%
{Francis \MakeLowercase{\textit{et al.}}: Long-Range Indoor Navigation with PRM-RL}
%



\maketitle

\begin{abstract}
Long-range indoor navigation requires guiding robots with noisy sensors and controls through cluttered environments along paths that span a variety of buildings. We achieve this with PRM-RL, a hierarchical robot navigation method in which reinforcement learning agents that map noisy sensors to robot controls learn to solve short-range obstacle avoidance tasks, and then sampling-based planners map where these agents can reliably navigate in simulation; these roadmaps and agents are then deployed on robots, guiding them along the shortest path where the agents are likely to succeed. Here we use Probabilistic Roadmaps (PRMs) as the sampling-based planner, and AutoRL as the reinforcement learning method in the indoor navigation context. We evaluate the method in simulation for kinematic differential drive and kinodynamic car-like robots in several environments, and on differential-drive robots at three physical sites. Our results show PRM-RL with AutoRL is more successful than several baselines, is robust to noise, and can guide robots over hundreds of meters in the face of noise and obstacles in both simulation and on robots, including over 5.8 kilometers of physical robot navigation.

\end{abstract}


%
\IEEEpeerreviewmaketitle

\section{Introduction}
\begin{figure}
\label{fig:navigation}
	\begin{center}
		\begin{tabular}{cc}
			\subfloat[Indoor navigation platform]{\includegraphics[trim=0mm 0mm 0mm 0mm,clip,width=0.22\textwidth,keepaspectratio=true]{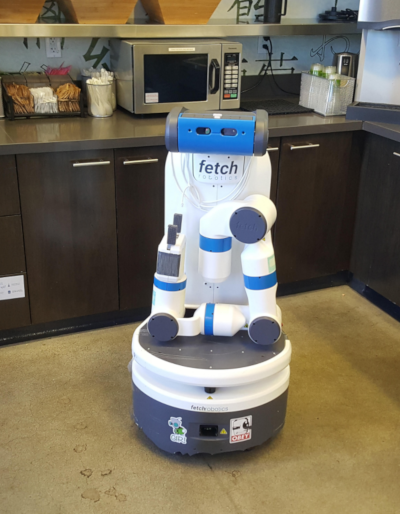}\label{fig:fetch}}
            &
            \subfloat[Physical Testbed 1]{\includegraphics[trim=0mm 0mm 0mm 6mm,clip,width=0.22\textwidth,keepaspectratio=true]{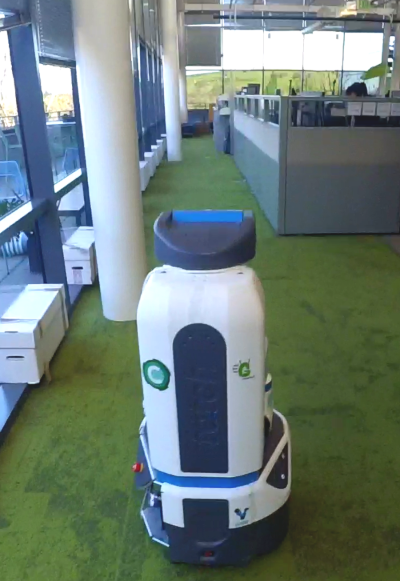}\label{fig:building}}            
			\\
		\end{tabular}
		\caption{\small The long-range indoor navigation task. (a) Approximately cylindrical differential drive robot. (b) Deployment environments are office buildings.
		}
	\end{center}
\end{figure}

\IEEEPARstart{L}{ong}-range indoor robot navigation requires human-scale robots (Fig. \ref{fig:fetch}) to move safely over building-scale distances (Figs. \ref{fig:building}). To robustly navigate long distances in novel environments, we factor the problem into long-range path planning and end-to-end local control, while assuming the robot has mapping and localization. Long-range path planning finds collision-free paths to distant goals not reachable by local control \cite{Lavalle06book}. End-to-end local control produces feasible controls to follow ideal paths while avoiding obstacles (e.g., \cite{khatib1986real}, \cite{fox1997dynamic}) and compensating for noisy sensors and localization  \cite{autorl}. We enable end-to-end local control to inform long-range path planning through sampling-based planning.

Sampling-based planners, such as Probabilistic Roadmaps (PRMs) \cite{kavraki-prm} and Rapidly Exploring Random Trees (RRTs)  \cite{kuffner-rrt,Lavalle-rrt2}, plan efficiently by approximating the topology of the configuration space ($\cspace$), the set of all possible robot poses, with a graph or tree constructed by sampling points in the collision-free subset of configuration space ($\cfree$), and connecting these points if there is a collision-free local path between them. Typically these local paths are created by line-of-sight tests or an inexpensive local planner, and are then connected in a sequence to form the full collision-free path.

\begin{figure*}[t]
	\begin{center}
		\begin{tabular}{c}
        \includegraphics[trim=0mm 0mm 0mm 0mm,clip,width=0.98\textwidth,keepaspectratio=true]{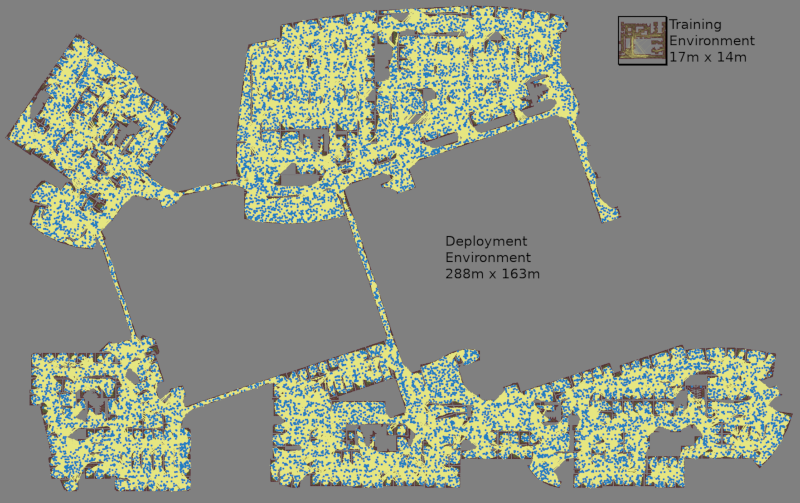}
		\end{tabular}
		\caption{\small Quad-Building Complex---\dist{288} by \dist{163}: A large roadmap derived from real building plans which PRM-RL successfully navigated $57.3\%$ of the time in simulation. The connected segment in the upper center corresponds to the upper floor of Building 1 used in our evaluations and contains the space where we collected our SLAM map in Fig.  \ref{fig:slam}. Blue dots are sampled points and yellow lines are roadmap connections navigable in simulation with the AutoRL policy. This roadmap has 15,900 samples and had 1.4 million candidate edges prior to connection attempts, of which 689,000 edges were added. It took 4 days to build using 300 workers in a cluster, requiring 1.1 billion collision checks. The upper right inset is the training environment from Fig. \ref{fig:mtv1965}, to scale; the quad-building complex is approximately two hundred times larger in map area.  \label{fig:googleplex}}
	\end{center}
\end{figure*}

\begin{figure*}[t]
	\begin{center}
		\begin{tabular}{ccc}
            \subfloat[Training environment---\dist{23} by \dist{18}]{\includegraphics[trim=0mm 0mm 0mm 0mm,clip,width=0.5\textwidth,height=4.5cm,keepaspectratio=true]{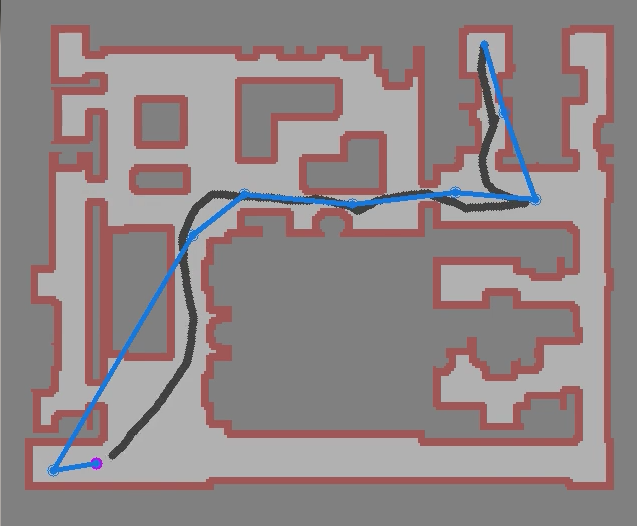}\label{fig:mtv1965}} 
			&
			\subfloat[Building 2---\dist{60} by \dist{47}]{\includegraphics[trim=0mm 0mm 0mm
			0mm,clip,width=0.5\textwidth,height=4.5cm,keepaspectratio=true]{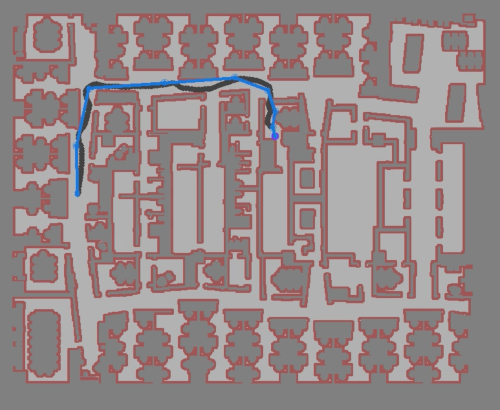}\label{fig:sfo}} 
			&
			\subfloat[Building 3---\dist{134} by \dist{93} ]{\includegraphics[trim=0mm 0mm 0mm 0mm,clip,width=0.5\textwidth,height=4.5cm,keepaspectratio=true]{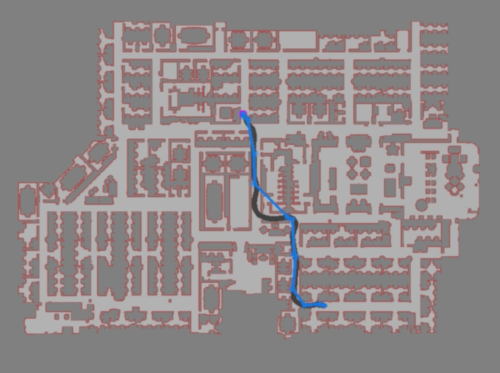}\label{fig:1667}}\\
			\multicolumn{3}{c}{\subfloat[Building 1---\dist{183} by \dist{66}]{\includegraphics[trim=0mm 12mm 0mm 5mm,clip,width=0.98\textwidth,keepaspectratio=true]{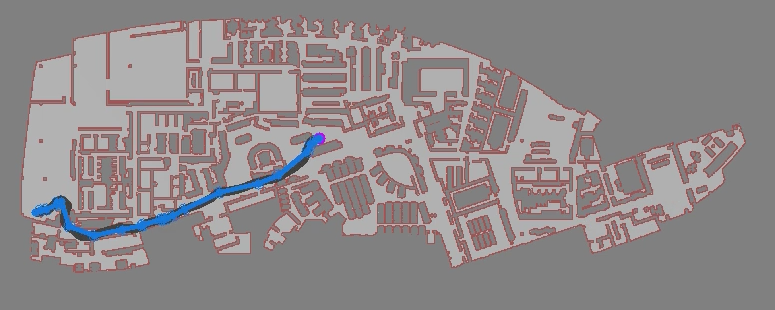}\label{fig:b40}}}
		\end{tabular}
		\caption{\small The environments used for indoor navigation are derived from real building plans. (a) The smallest environment is used to train the RL agent for faster training and iteration. (b)-(d) PRMs for deployment environments are built using agents trained in the training environment. Red \added{regions are}{fields are regions} deemed too close to obstacles \added{and}{which} cause episode termination when the robot enters them; light grey is free space from which the starts and goals are selected. Blue lines connect PRM waypoints, and the RL agent's executed trajectory is black. \label{fig:maps}}
	\end{center}
\end{figure*}

Regardless of how a planner generates a path, executing a path requires handling sensor noise, unmodeled dynamics, and environment changes. Recently, reinforcement learning (RL) agents \cite{jan-peters-ijrr-survey} have solved complex robot control problems \cite{DBLP:journals/corr/YahyaLKCL16}, generated trajectories under task constraints \cite{faust-ai-journal}, demonstrated robustness to noise \cite{faust-icra-15}, and learned complex skills \cite{bartt-mkp-11} \cite{atari-paper}, making them good choices to deal with task constraints. Many simple navigation tasks only require low-dimensional sensors and controls, like lidar and differential drive, and  can be solved with easily trainable networks \cite{zhu2016target,DBLP:journals/corr/GuptaDLSM17,brahmbhatt2017deepnav}.
However, as we increase the problem's complexity by requiring longer episodes or providing only sparse rewards \cite{pearl}, RL agents become harder to train and do not consistently transfer well to new environments \cite{kaelbling1996reinforcement} \cite{irpan-rl-hard}. 

Long-range navigation presents all these challenges. Sparse rewards and long episodes make long-range agents hard to train. On complex maps, short-range agents are vulnerable to local minima like wide barriers and narrow passages. Even within deployment categories, environments have vast variety: the SunCG dataset had 45,000 houselike environments \cite{song2016ssc}, and the US alone has over 5.6 million office buildings \cite{cbecs-2012}.

We present PRM-RL, an approach to long-range navigation which combines PRMs and RL to overcome each other's shortfalls. In PRM-RL, an RL agent learns a local point-to-point task, incorporating system dynamics and sensor noise independent of long-range environment structure. The agent's learned behavior then influences roadmap construction; PRM-RL builds a roadmap by connecting two workspace points only if the agent consistently navigates between them in configuration space without collision, thereby learning the long-range environment structure. PRM-RL roadmaps perform better than roadmaps based on pure $\cfree$ connectivity because they respect robot dynamics. RL agents perform better with roadmap guidance, avoiding local minima. PRM-RL thus combines PRM efficiency with RL resilience, creating a long-range navigation planner that not only avoids local-minima traps, but transfers well to new environments, as shown in Fig.  \ref{fig:googleplex}, where a policy trained on a small training environment scales to a quad-building complex almost two hundred times larger in map area. 

This paper contributes PRM-RL as a hierarchical kinodynamic planner for navigation in large environments for robots with noisy sensors. This paper is a journal extension of our conference paper \cite{prm-rl}, which contributes the original PRM-RL method. Here, we investigate PRM-RL in the navigation context and make the following contributions beyond the original paper:
1) Algorithm \ref{alg:prm-rl} for PRM-RL roadmap building;
2) Algorithm \ref{alg:navigate} for robot deployment;
3) PRM-RL application to kinodynamic planning on a car model with inertia; 
4) in-depth analysis of PRM-RL, including: 
4.1) correlation between the quality of the local planner and the overall hierarchical planner;
4.2) impact of improving planners and changing parameters on PRM-RL computational time complexity;
4.3) impact of a robust local planner on the effective connectivity of samples in the graph;
4.4) resilience of PRM-RL to noise and dynamic obstacles.
All the evaluations and experiments are new and original to this paper. We evaluate PRM-RL with a more effective local planner \cite{autorl}, compare it in simulation against six baselines in eight different buildings, and deploy it to three physical robot testbed environments. 

Overall, we show improved performance over both baselines and our prior work, more successful roadmaps, and easier transfer from simulation to robots, including a $37.5\%$ increase in navigation success over \cite{prm-rl}, while maintaining good performance despite increasing noise. We also show that only adding edges when agents can always navigate them makes roadmaps cheaper to build and improves navigation success; denser roadmaps also have higher simulated success rates, but at substantial roadmap construction cost. Floorplans are not always available or up to date, but we show roadmaps built from SLAM maps close the simulation to reality gap by producing planners which perform almost as well on robot as they do in simulation. SLAM-based PRM-RL enables real robot deployments with up to +\dist{200} collision-free trajectories at three different sites on two different robots with success rates as high as $92.0\%$. We also show that PRM-RL functions well on robots with dynamic constraints, with an $83.4\%$ success rate in simulation.

While this paper focuses on navigation, the analysis and empirical findings will be of interest to the wider motion planning community for two reasons. First, PRM-RL presented here is an example of a hierarchical motion planner that factors models of sensor, localization, and control uncertainties into roadmap construction, resulting in planners that perform as well in simulation as they do on robots. Second, we present a comprehensive analysis of the trade-offs between performance and computational complexity and the interplay between local and global planners that is not specific to navigation.

\section{Related Work}
\label{sec:related-work}
\subsubsection{Probabilistic roadmaps} PRMs \cite{kavraki-prm} have been used in a wide variety of planning problems from robotics \cite{latombe-lunar,malone-embodied2012} to molecular folding \cite{AlBluwi_survey_2012,park-prmrl,tapia_motion_2010}.
They have also been integrated with reinforcement learning for state space dimensionality reduction \cite{malone-journal-14,park-prmrl} by using PRM nodes as the state space for the reinforcement learning agent. 
In contrast, our work applies reinforcement learning to the full state space as a local planner for PRMs. 
In prior work for an aerial cargo delivery task, we trained RL agents to track paths generated from PRMs constructed using a straight-line local planner \cite{faust-ai-journal}.
Researchers have modified PRMs to handle moving obstacles \cite{hsu-moving-02, amato-moving-07}, noisy sensors  \cite{malone-iros-2013}, and localization errors \cite{amato-uncertain-13, alterovitz-rss-07}.
Safety PRM \cite{malone-iros-2013} uses probabilistic collision checking with a straight-line planner, associating a measure of potential collision with all nodes and edges.
All those methods address one source of errors at a time. In contrast, PRM-RL uses an RL-based local planner capable of avoiding obstacles and handling noisy sensors and dynamics, and at the node connection phase the RL local planner does Monte Carlo path rollouts with deterministic collision checking.
We only add edges if the path can be consistently navigated within a tunable success threshold. Additionally, PRMs built with straight-line geometric planners require a tracking method (often model-based, such as \cite{lqg-mp}) to compensate for sensor and dynamics uncertainties not known at the planning time. PRM-RL eliminates the need for path tracking and mathematical models for sensor and dynamics uncertainties, because sensor-to-control RL local planners are model-free, and the Monte Carlo rollouts ensure that the feasibility is accounted when connecting the nodes.

PRMs are easy to parallelize, either through parallel edge connections \cite{nancy-parallel}, sampling \cite{ron-parallel}, or building sub-regional roadmaps \cite{neighbor-connections} in parallel. To speed up building large scale roadmaps, we use an approach similar to \cite{nancy-parallel} across different computers in a cluster. Individual Monte Carlo rollouts to connect edges can be parallelized across multiple processors or run sequentially to allow for early termination.

\subsubsection{Reinforcement learning in motion planning} Reinforcement learning has recently gained popularity in solving motion planning problems for systems with unknown dynamics \cite{jan-peters-ijrr-survey}, and has enabled robots to learn tasks that have been previously difficult or impossible \cite{abel2016exploratory,chen2015deepdriving,levine2016end}. For example, Deep Deterministic Policy Gradient (DDPG) \cite{ddpg} works with high dimensional continuous state/action spaces and can learn to control robots using unprocessed sensor observations \cite{levine2016end}.

Researchers have successfully applied deep RL to navigation for robots, including visual navigation with simplified navigation controllers \cite{brahmbhatt2017deepnav,duan2016rl,DBLP:journals/corr/GuptaDLSM17,mousavi_eccv18,seff2016learning,zhu2017target}, more realistic controllers in game-like environments \cite{bhatti2016playing,dosovitskiy2016learning,mirowski2016learning}, and extracting navigation features from realistic environments \cite{chen2015deepdriving,giusti2016machine}. In local planner settings similar to ours, differential drive robots with 2D lidar sensing, several approaches emerged recently using asynchronous DDPG \cite{virtualtai2017}, expert demonstrations \cite{Pfeiffer2017FromPT}, DDPG \cite{lidar-rl}, curriculum learning \cite{successor-features}, and AutoRL \cite{autorl}. While any sensor-to-controls obstacle-avoidance agent could be used as both a local planner in PRM-RL and a controller for reactive obstacle avoidance, we choose AutoRL for its simplicity of training, as it automates the search for reward and network architecture. 

Recent works \cite{optimal-q,successor-features} learn planning in 2D navigation mazes and measure transferability of learning to new environments. Using the terminology of this paper, this work falls between long-range planning and local control. Their navigation environments vary between task instances, as in both our AutoRL and PRM-RL setting. The mazes appear to be similar in size and complexity to the maps used in our local planning \cite{autorl,prm-rl}. However, those methods, based on Q-learning, use discretized actions to produce their paths, leading to three consequences. First, the robot's action discretization inherently limits how close to the optimal path the solution can reach. Second, the method must have a lower-level controller or steering function that tracks the resulting path. Third, while these planners avoid obstacles, they do not take robot dynamics or path feasibility into account. In our work, we rely on continuous action RL algorithms, which can better approximate optimal paths than discrete actions, eliminate the tracking controller, and learn dynamically feasible steering functions that produce linear and angular velocities \cite{rlrrt}.

\subsubsection{Hierarchical planners with reinforcement learning} Several recent works feature hierarchical planners combined with reinforcement learning, either over a grid \cite{unstuck-dinesh} or manually selected waypoints \cite{ddqn-topological}. These works connect roadmap points with a straight-line planner and use the reinforcement learning agent as an execution policy at run-time. We use the obstacle-avoidance RL policy as both an execution policy and a local planner for connecting the edges in the roadmap. This approach results in roadmap connectivity that is tuned to the abilities of the particular robot.

\section{Problem Statement}
\label{sec:problem}
This section defines key terms for path planning, for trajectory planning, for the navigation problem for robots with noisy depth sensors and actuators, and for the differential drive and car-like robots which are our primary focus in this paper.

The \textit{configuration space}, $\cspace,$ is the set of possible robot \textit{poses}. The configuration spaces for differential drive and car-like robots are $\cspace_{dd} = \mathbb{R}^2 \times S^1$ and $\cspace_{car} = \mathbb{R}^2 \times S^1 \times S^1,$ respectively. The \textit{workspace}, $\mathcal{W}$, is the physical space that a robot operates in with dimensionality $D_\mathcal{W}$ of 2 or 3. The workspace and the robot's kinematics divide the configuration space into valid ($\cfree$) and invalid partitions. $\cfree$ is a set of all robot poses that are free of self-collision, collisions in the workspace, and satisfy relevant kinodynamic constraints. To that end, we consider the workspace to be a closed segment on a 2-dimensional manifold, and model the robots' kinematics with a unicycle or Type (2,0) model \cite{siciliano2016springer} for differential drive robots, and single-track model \cite{DBLP:journals/corr/PadenCYYF16} for car-like robots.

The robot \textit{state space} is the full internal state of the robot including pose, velocity, observations, etc. We assume this state to be hidden and non-observable. The observable state space, $O \subset \R{N_s\theta_n} \times \cfree^2,$ is the same for both robot types and consists of sensor observations ($N_s$ lidar rays observed over the last $\theta_n$ discrete time steps) as well as the current and goal robot poses, assumed to be in $\cfree.$ The robot \textit{action space}, $A \subset \R{2}$, consists of linear and angular velocities $\ac = (v, \phi) \in A$ for both types of robots. We assume sensor observation and actuators to be noisy. The sensor observations are produced by a sensor process $F_{s} : \cfree \rightarrow O$  that can be modeled as a combination of inherent sensor dynamics and a source of noise: $F_{s}(\myvec{x}) \sim D_{s}(\myvec{x}) + \mathcal{N}_{s}$. Similarly, actions in the robot's action space $A$ have a state-dependent effect $F_{a} : \cfree \times A \rightarrow \cspace$, which also can be modeled as a combination of inherent robot dynamics and a source of noise: $F_{a}(\myvec{x}, \myvec{a}) \sim D_{a}(\myvec{x}, \myvec{a}) + \mathcal{N}_{a}$.

A \textit{path}, $\mathcal{P},$ is a sequence of workspace points $\myvec{p_i} \in \mathcal{W}, i \in [0, N]$ from the beginning $p_0$ to the end $p_N$ of the trajectory. A \textit{valid path} consists of only valid waypoints: $\forall \myvec{p_i} \in \mathcal{P} : \myvec{p_i} \in \mathcal{W} \cap \cfree, i=1,\cdots,N_p$. We consider a \textit{trajectory,} $\mathcal{T}$, to be a sequence of robot valid poses $\myvec{x}_j \in \cfree,\, j=1,\cdots,N_t$ such that $\myvec{x}_{j+1}$ is reachable from $\myvec{x}_j$ within $\cfree$ under the robot's kinematic model within a fixed discrete time step $\Delta T,$ for any two consecutive points $\myvec{x}_{j+1},\, \myvec{x}_j,\, j=1,\cdots,N_t-1$. We assume that the robot is at rest at the beginning and end of the trajectory, i.e., $\dot{\myvec{x}}_1 = \myvec{0},\, \dot{\myvec{x}}_N = \myvec{0}.$

In this paper, a \textit{point-to-point (P2P)} policy $\myvec{\pi}:O \rightarrow A,$ maps robot observations to linear and angular velocities in order to generate trajectories. 
Given a valid start configuration $\myvec{x}_{S}$ and a policy $\myvec{\pi}$, an \textit{executed trajectory} $\mathcal{T}$ is a sequence of configuration states that result from drawing actions from the policy and its noise processes: $\mathcal{T} : \myvec{x_0} = \myvec{x_{S}} \land \myvec{x_{i}} \sim  F_{a}(\myvec{x_{i-1}}, \myvec{\pi}(F_{s}(\myvec{x_{i-1}})))$. An executed trajectory is a \textit{failure} if it produces a point that exits $\cfree$, or if it exceeds a task-specific time-step limit $\mathcal{K}_{\omega}$ without reaching a goal. Given a valid observable goal pose, $\myvec{x_{G}},$ a non-failed trajectory is a \textit{success} if it reaches a point $\myvec{x_{i}}$  sufficiently close to the goal with respect to a task-dependent threshold $\|\myvec{x_{i}}-\myvec{x_{G}}\| \leq d_{G}$, at which point the task has \textit{completed}.

Graph search over PRMs creates a path, $\mathcal{P}$. Waypoints in the path serve as intermediate goals for the trajectory generating policy $\myvec{\pi}.$ 
A \textit{path following policy} with respect to a path $\mathcal{P}$ and P2P policy $\myvec{\pi},$ $\myvec{\pi}_{\textit{pf}}(\myvec{x} | {\mathcal{P}}, \myvec{\pi}),$ produces a trajectory that traverses the waypoints in path $\mathcal{P}$ using the P2P policy $\myvec{\pi}.$ A \textit{valid executable path} with respect to a P2P policy is a path which the P2P policy can reliably execute to achieve task completion---guiding the agent from the start state $\myvec{x_{S}}$ of $\mathcal{P}$ to  within $d_{G}$ of the goal state $\myvec{x_{G}}$ within $\mathcal{K}_{\omega}$ time steps. 
Because noise makes execution stochastic, we define a path to be \textit{reliable} if the policy's probability of task completion using the path exceeds a task-dependent \textit{success threshold} $p_{s}$.

The key problem that PRM-RL addresses is how to construct a \textit{reliable path} and \textit{path following policy} for a given P2P policy in the context of indoor navigation. To that end, we define an agent that performs its task without knowledge of the workspace topology as one in which the transfer function $\dot{\myvec{x}} = f(\myvec{x}, \uv)$ that leads the system to task completion is only conditioned on what the robot can observe and what it has been commanded to do. Formally, we learn policies to control an agent that we model as a partially observable Markov decision process (POMDP) represented as a tuple $(O, A, D, N, R, \gamma)$ of observations, actions, dynamics, noise, reward, and discount. The characteristics of the robot determine observations, actions, dynamics, and noise; these are continuous and multi-dimensional. Reward and discount are determined by the requirements of the task: $\gamma \in (0,1) $ is a scalar discount factor, whereas the reward $R$ has a more complicated structure $(G, r)$, including a true scalar objective $G$ representing the task and a weighted dense scalar reward $r:O \times \R{N_\theta} \rightarrow \R{},$ based on observable features $O$ and a reward parameterization $\theta \in \R{N_\theta}$. We assume a presence of a simplified black-box simulator without knowing the full non-linear system dynamics. The dynamics $D$ and noise $N$ are implicit in the real world but are encoded separately in simulation in the system dynamics and an added noise model.

\begin{figure}[t]
	\begin{center}
		\begin{tabular}{c}
            \subfloat[RL Training]{\includegraphics[scale=0.5,width=0.48\textwidth,height=4.cm]{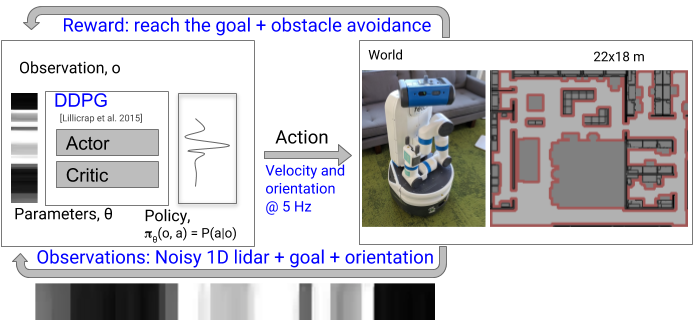}\label{fig:prm-flowchart}} 
			\\
            \subfloat[PRM-RL Deployment]{\includegraphics[scale=0.5,width=0.48\textwidth,height=8cm]{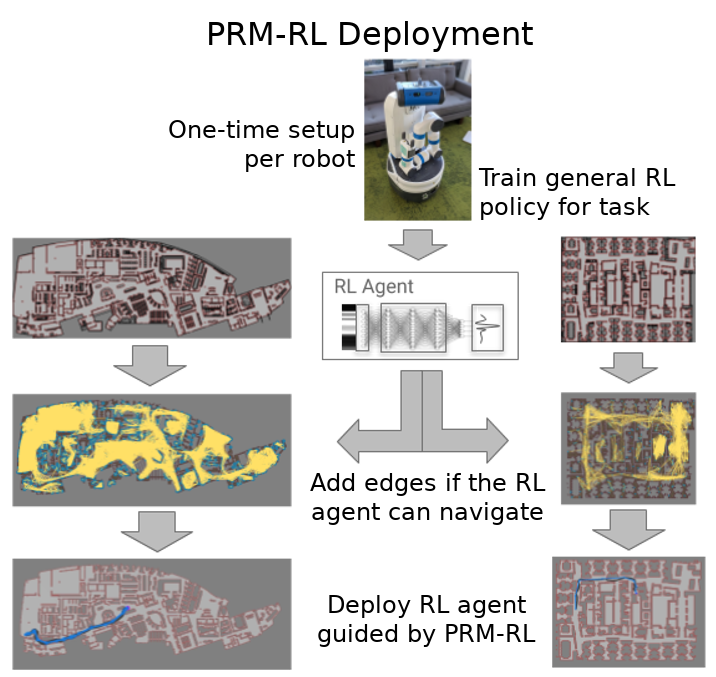}\label{fig:prm-deployment}}\\
		\end{tabular}
		\caption{\small The PRM-RL approach. (a) RL learns a model of task and system dynamics. This enables the construction of a local planner and the generation of a PRM-RL roadmap. This roadmap and policy can then be executed on the robot using the same local planner. (b) The same policy can generate roadmaps for different floorplans, enabling deployment to many sites.\label{fig:arch}}
	\end{center}
\end{figure}

\section{Methods}
\label{sec:methods}
The PRM-RL method has three stages: training an environment-independent local planner policy with RL, creating a roadmap specific to that local planner and an environment, and deploying that roadmap and policy to the environment for querying, trajectory generation, and execution. Fig. \ref{fig:arch} illustrates the method.

First, in the training stage (Fig. \ref{fig:prm-flowchart}), to enable a robot to perform a specific task, we train an agent with RL. For indoor navigation, that task is short-range point-to-point navigation end-to-end from sensors to actions. This task is independent of the environment in which the robot will eventually operate. The RL agent learns to perform a task on an environment comparable to the deployment environment, but smaller in size to make simulation faster and training more tractable. This is a Monte Carlo simulation process: we train multiple policies and select the fittest one for the next stage of PRM-RL, regardless of the learning algorithm used.

Next, in the creation phase (Fig. \ref{fig:prm-deployment} upper), to prepare a robot to work in a specific environment, we use this best policy as a local planner to build a PRM with respect to the target site. Obstacle maps, such as floor plans or SLAM maps, can be used for any robot we wish to deploy as long as the policy transfers well to the real world. This is a one-time setup per robot and environment. Unlike PRM-SL, in which points are sampled from $\cfree$, PRM-RL samples points from the workspace $\mathcal{W}$ but then rolls out trajectories in $\cspace$, adding an edge to the roadmap only when the agent can navigate it in simulation with greater probability than $p_{s}$ over $n_{\omega}$ trials. Rather than being determined by the geometry of free space, the resulting roadmap is tuned to capabilities of the particular robot, so different robots over the same floor plan may generate different roadmaps with different connectivity.

Finally, in the deployment phase, (Fig. \ref{fig:prm-deployment} lower), to perform the task in the environment, the constructed roadmap is queried to generate trajectories, which are executed by the same RL agent used to generate the roadmap. In the geometric PRM-SL framework, querying a roadmap  involves connecting the start and goal to the roadmap graph and finding the shortest path between them. In the simulation-based PRM-RL framework we can optionally record additional data about executed trajectories to enable other trajectory metrics (such as minimal energy consumption, shortest time, and so on) that are not generally available with geometry-only approaches. At execution time, the RL agent navigates to the first waypoint. Once the agent is within $d_{G}$ distance from the waypoint, the next waypoint becomes its new goal; the process repeats until the agent has traversed the whole trajectory.

\subsection{RL Agent Training}
\label{sec:rl-agent-training}

PRM-RL's global planner is strongly decoupled from the details of the local planner's construction and training. We explore this with two different agent models: differential drive and carlike robots.

\subsubsection{P2P for differential drive robots}
\label{sec:p2p-diff-drive}
The true objective of the P2P agent is to maximize the probability of reaching the goal without collisions,
\begin{equation}
\label{eq:p2p_obj}
G_{\text{P2P}}(\myvec{x_{i}}, \myvec{x_{G}}) = \mathbb{I} (\|\myvec{x_{i}} - \myvec{x_{G}} \| < d_{G}),
\end{equation}
where $\mathbb{I}$ is an indicator function, $\myvec{x_{G}}$ is the goal position, and $d_{G}$ is the goal radius. The zero-collision property is enforced by terminating episodes on collisions. The goal observation $\myvec{o}_g$ is the relative goal position in polar coordinates, which is readily available from localization. The reward for P2P for differential drive robots is the dot product of the parameters and reward components:
\begin{equation}
R_{\myvec{\theta}_{r_\text{DD}}} = 
\myvec{\theta}^{^\intercal}_{r_\text{DD}} [
r_\text{goal} \,
r_\text{goalDist} \,
r_\text{collision} \,
r_\text{clearance} \,
r_\text{step} \,
r_\text{turning} \,
]^\intercal,
\label{eq:p2p_reward}
\end{equation}
where
$r_\text{goal}$ is 1 when the agent reaches the goal and 0 otherwise,
$r_\text{goalDist}$ is the negative Euclidean distance to the goal,
$r_\text{collision}$ is 1 when the agent collides with obstacles and 0 otherwise,
$r_\text{clearance}$ is the distance to the closest obstacle,
$r_\text{step}$ is a constant penalty step with value 1, and
$r_\text{turning}$ is the negative angular speed.
We train this model with AutoRL \cite{autorl} over the DDPG \cite{ddpg} algorithm, which simultaneously finds the reward weights $\myvec{\theta_{r_\text{DD}}}$ and trains the agent. AutoRL automates hyperparameter search in reinforcement learning using an evolutionary approach. AutoRL takes as inputs a true objective used to evaluate the agent, a parameterized dense reward that the agent uses to train itself, and optionally neural network architecture and algorithm hyperparameters. To train the agent, AutoRL typically optimizes these hyperparameters in phases. First, given an arbitrary or hand-tuned architecture, it trains several populations of RL agents with different reward parameters and optimizes over the true objective. Optionally, a second phase repeats the process with the dense reward fixed while searching over neural network architectures instead.

\subsubsection{P2P for carlike robots}
\label{sec:p2p-car-model}
The true objective of P2P does not change for car drive, but because the turn radius of the car is limited and the car must perform more complex maneuvers, we choose a slightly different reward model:
\begin{equation}
R_{\myvec{\theta}_{r_\text{CM}}} = 
\myvec{\theta}^{^\intercal}_{r_\text{CM}} [
r_\text{goal} \,
r_\text{goalProg} \,
r_\text{collision} \,
r_\text{step} \,
r_\text{backward} \,
]{^\intercal},
\label{eq:p2p_car_reward}
\end{equation}
where all values are the same as for diff drive except 
$r_\text{goalProg}$ rewards the delta change of Euclidean distance to goal, and
$r_\text{backwards}$ is the negative of backwards speed and 0 for forward speed. We dropped $r_\text{goalDist}$, $r_\text{clearance}$, and $r_\text{turning}$ to reduce the space of hyperparameter optimization based on analysis of which parameters seemed to have the most positive impact upon learning differential drive models.
We train this model with hyperparameter tuning with Vizier \cite{vizier} over the DDPG \cite{ddpg} algorithm in a different training regime in which the car model is allowed to collide up to 10 times in training, but is still evaluated on the true objective of zero collisions.

\begin{algorithm}[t]
	\caption{PRM-RL AddEdge} 
	\label{alg:add_edge}
\begin{algorithmic}[1]
\small
\INPUT $s,\,g \in \mathcal{W} \cap \cfree$: Start and goal in workspace.
\INPUT $p_{s} \in [0, 1]$ Success threshold.
\INPUT $n_{\omega}$: Number of attempts.
\INPUT $d_{G}$: Sufficient distance to the goal.
\INPUT $\mathcal{K}_{\omega}$: Maximum steps for trajectory.
\INPUT $L(\x)$: Task predicate.
\INPUT $\pi$: RL agent's policy.
\INPUT $D$ Generative model of system dynamics.
\OUTPUT $add_{edge}, success_{rate}$ , $length$
\STATE $success_\text{rate} \leftarrow 0, failure \leftarrow 0, length \leftarrow 0$
\FOR {$i=1,\cdots n_{\omega}$ }
  \STATE $ $ // Run in parallel, or sequential for early termination.
  \STATE $\x_s \leftarrow s.\text{SampleConfigSpace}()$ // Sample from the
  \STATE $\x_g \leftarrow g.\text{SampleConfigSpace}()$ // $\cfree$ space.
  \STATE $success \leftarrow 0$, $steps \gets 0$, $\x \gets \x_s$
  \STATE $length_\text{trial} \leftarrow 0$
  \WHILE {$steps < \mathcal{K}_{\omega} \wedge 
          \|p(\x)-p(\x_g)\| > d_{G} \wedge p(\x) \in \cfree$  }
    \STATE $\x_p \leftarrow \x$, $\uv \leftarrow \pi(\x)$
	\STATE $\x \leftarrow D.predictState(\x, \uv)$
    \STATE $steps \leftarrow steps + 1$
    \STATE $length_\text{trial} \leftarrow length_\text{trial} + \|\x - \x_p \|$
  \ENDWHILE
  \IF {$\|p(\x)-p(\x_g)\| < d_{G}$}
    \STATE $success \leftarrow success + 1$
    \STATE $length_{trial} \leftarrow length_\text{trial} + \| p(\x) - p(\g) \|$
  \ELSE
    \STATE $failure \leftarrow failure + 1$
  \ENDIF
  \IF {$(1 - p_{s}) < failure / n_{\omega}$}
    \RETURN False, 0, 0\added{}{, 0} // Not enough success, we can terminate.
  \ENDIF
  \STATE $length \leftarrow length + length_{trial}$
\ENDFOR
\STATE $length \leftarrow \frac{length}{success}$, $success_\text{rate} \leftarrow \frac{success}{n_{\omega}}$
\RETURN  $success_{rate} > p_{s}, success_{rate}, length$
\end{algorithmic}
\end{algorithm}

\subsection{PRM Construction}
\label{ss:mp}

The basic PRM method works by sampling robot configurations in the the robot's configuration space, retaining only collision-free samples as nodes in the roadmap. PRMs then attempt to connect the samples to their nearest neighbors using a local planner. If an obstacle-free path between nodes exists, PRMs add an edge to the roadmap.

We modify the basic PRM by changing the way nodes are connected. Formally, we represent PRMs with graphs modeled as a tuple $(V, E)$ of nodes and edges. Nodes are always in free space, $V_i \in \cfree$, and edges always connect valid nodes $(V_i, V_j)$, but we do not require that the line of sight $\overline{V_i V_j}$ between those nodes is in $\cfree$, allowing edges that ``go around'' corners. Since we are primarily interested in robustness to noise and adherence to the task, we only connect configurations if the RL agent can consistently perform the point-to-point task between two points. 

Algorithm \ref{alg:add_edge} describes how PRM-RL adds edges to the PRMs. We sample multiple points from the configuration space, around the start and goal in the workspace, and attempt to connect the two points over $n_{\omega}$ trials. An attempt is successful only if the agent reaches sufficiently close to the goal point. Even with a high success threshold, PRM-RL trajectories are not guaranteed to be collision-free because of sensor and action noise. To compute the total length of a trajectory, we sum the distances for all steps plus the remaining distance to the goal. The length we associate with the edge is the average of the distance of successful edges. The algorithm adds the edge to the roadmap if the success probability $success_{rate}$ is above the threshold $p_{s}$.  

The worst-case number of collision checks in Algorithm \ref{alg:add_edge} is $O(\mathcal{K}_{\omega} * n_{\omega})$, because multiple attempts are required for each edge to determine $success_{rate}$. Each trial of checking the trajectory can be parallelized with $n_{\omega}$ processors; alternately, trajectory checking within Algorithm \ref{alg:add_edge} can be serialized, terminating early if the tests fail too many times. Mathematically, for a given success threshold and desired number of attempts, at least $n_{s} = \ceil{p_{s} * n_{\omega}}$ trials must succeed; therefore we can terminate when $n_{s} > p_{s} * n_{\omega}$, or when the failures exceed the complementary probability $n_{f} > (1-p_{s}) * n_{\omega}$. This can provide substantial savings if $p_{s}$ is high, as shown in Section \ref{sec:results-connect}. Much of PRM-RL construction can be parallelized; parallel calls to Algorithm \ref{alg:add_edge} can speed roadmap construction, and roadmaps can be constructed in parallel.

We use a custom kinematic 3D simulator which provides drive models for agents and supports visual sensors such as cameras and lidars. The simulator also provides parameterized noise models for actions, observations, and robot localization, which improves model training and robustness. Stepping is fast compared to full-physics simulations because our simulator is kinematic. This speeds up RL training and roadmap building.

\begin{algorithm}[t]
	\caption{PRM-RL Build roadmaps} 
	\label{alg:prm-rl}
\begin{algorithmic}[1]
\small
\INPUT Obstacle maps: $[m1, ..,m_n]$ 
\INPUT $\pi$: RL agent's policy.
\INPUT $D$: Generative model of system dynamics.
\INPUT $\rho_{\omega}$: Sampling density.
\INPUT $d_{\pi}$: Policy range.
\INPUT $n_p$: Number of processors.
\INPUT $p_{s} \in [0, 1]$: Success threshold.
\INPUT $n_{\omega}$: Number of attempts.
\OUTPUT RL policy, $\pi,$ Roadmaps, $[roadmap_1,.., roadmap_n]$
\STATE Train RL agent with \cite{autorl} given $D$ as described in Section \ref{sec:rl-agent-training}.
\FOR {$m$ in $[m_1, ..,m_n]$ /* In parallel for each env. */}
  \STATE Sample environment map $m$ with density $\rho_{\omega}$ and store candidate edges \wrt $d_{\pi}$.
  \STATE Partition candidate edges in $n_p$ subsets, $[e_1, ..., e_{n_p}]$.
  \FOR {$edges$ in $[e_1, ..., e_{n_p}]$ /* In parallel over workers. */}
    \FOR {$e$ in $edges$ /* In parallel over threads. */}
      \IF {AddEdge: Run Alg \ref{alg:add_edge} with $\pi.$}
        \STATE Add nodes if not in $roadmap$.
        \STATE Add edge $e$ to the $roadmap$.
      \ENDIF
    \ENDFOR
  \ENDFOR
\ENDFOR
\RETURN RL policy, $\pi,$ Roadmaps, $[roadmap_1,.., roadmap_n]$
\end{algorithmic}
\end{algorithm}

Algorithm \ref{alg:prm-rl} describes the roadmap building procedure, where an RL agent is trained once, and used on several environments. While building a roadmap for each environment, we first sample the obstacle map to the given density and store all candidate edges. Two nodes that are within the RL policy range, $d_{\pi},$ are considered candidates. Next, we partition all the candidate edges into subsets for distributed processing. The PRM builder considers each candidate edge and adds it, along with its nodes, to the roadmap if the AddEdge Monte Carlo rollouts returns success above threshold.

\begin{algorithm}[t]
	\caption{PRM-RL Navigate} 
	\label{alg:navigate}
\begin{algorithmic}[1]
\small
\INPUT PRM $roadmap.$
\INPUT $\pi$: RL agent's policy.
\INPUT $s,\,g \in \cfree$: Start and goal.
\INPUT $d_{G}$: Sufficient distance to the goal.
\INPUT $\mathcal{K}_{\omega}$: Maximum steps for trajectory.
\STATE Add start and goal $s,\,g \in \cfree,$ to $roadmap$ if needed.
\STATE Query $roadmap$ and receive list of waypoints $[w_1,..,w_N],\, w_1 = s,\,w_N = g$.
\FOR {$w$ in $[w_2,..,w_N],$}
  \STATE Set $w$ as a goal for the RL agent $\pi$.
  \STATE Set current state, $c$ as start state $s.$
  \STATE ${steps} \leftarrow 0$
  \WHILE {$c$ is not within ${d_{G}}$ from $w$}
    \STATE Apply action $\pi(c),$ and observe the resulting state as new current state $c.$
    \STATE $steps \leftarrow steps + 1$
    \IF {$c$ is in collision}
      \RETURN Error: Collision.
    \ENDIF
    \IF {$steps > \mathcal{K}_{\omega}$}
      \RETURN Error: Timeout.
    \ENDIF
  \ENDWHILE
\ENDFOR
\RETURN Success.
\end{algorithmic}
\end{algorithm}

\subsection{Navigation}
\label{sec:navigation}
Finally, Algorithm \ref{alg:navigate} describes the navigation procedure, which takes a start and a goal position. These are added to the roadmap if not present. Then, the roadmap is queried for a list of waypoints. If no waypoints are returned, the algorithm returns the start and goal as the path to give the RL agent the opportunity to attempt to navigate on its own. In execution, a higher-level PRM agent gives the RL agent one waypoint at the time as a sub-goal, clearing these goals sequentially as the RL agent gets within goal distance $d_{G}$, until the final destination is reached or $\mathcal{K}_{\omega}$ is exceeded.  

\section{Results}
\label{sec:results}
We evaluate PRM-RL's performance on both floorplan maps and SLAM maps with respect to comparable baselines, construction parameters, simulated noise, and dynamic obstacles, as well as with experiments on robots. Section \ref{sed:methodology} describes the robot and training setup, evaluation environments, and baselines. Section \ref{sec:results-floorplan} demonstrates PRM-RL's superior performance on floormaps with respect to baselines, while the following sections examine PRM-RL's characteristics in more depth. Section \ref{sec:results-noise} demonstrates PRM-RL's robustness to noise, and Section \ref{sec:results-connect} explores quality and cost tradeoffs with success threshold and sampling density. Since one of our goals is to assess PRM-RL for real-world navigation, Sections \ref{sec:results-slam} and \ref{sec:results-large} show PRM-RL's applicability to SLAM and large-scale maps, and  Section \ref{sec:results-robots} analyzes sim2real experiments on real robots. Finally, to demonstrate the extensibility of PRM-RL to new situations,  Section \ref{sec:results-car} shows PRM-RL works well on simulated robots with dynamic constraints, Section \ref{sec:results-dynamic} explores PRM-RL's robustness in the face of dynamic obstacles, and Section \ref{sec:results-synthetic} compares PRM-RL's performance on a suite of maps used by a visual policy baseline.

\subsection{Methodology}
\label{sed:methodology}
\subsubsection{Robot setup}
\begin{figure}[tb]
\centering
\includegraphics[width=0.48\textwidth,keepaspectratio=false]{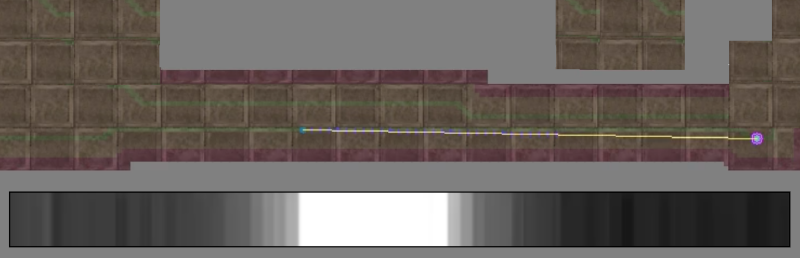}
\caption{\small Lidar observation that the robot uses for navigation. The background image shows a hallway with a clear path ahead of the robot and walls to the left and a right. The inset shows the lidar image: white regions indicate no obstacles within \dist{5}, and progressively darker regions indicate closer and closer obstacles.\label{fig:lidar}}
\end{figure}

We use two different robot kinematic models, \textit{differential drive} \cite{Lavalle06book} and \textit{simple car model} \cite{Lavalle06book, DBLP:journals/corr/PadenCYYF16}. We control both models with linear and angular velocities commanded at $5$ Hz, receive goal observations from off-the-shelf localization, and represent both as circles with \dist{0.3} radius. The obstacle observations are from 2D lidar data (Fig. \ref{fig:lidar}), with a $220\degree$ FOV resampled to $64$ rays. Following \cite{atari-paper}, we use an observation trace of the last $\theta_n$ frames to provide a simple notion of time to the agent, with $\theta_{n_{CM}} = 3$ and $\theta_{n_{DD}} ]= 1$ for the car model and diff-drive respectively. We use Fetch robots \cite{fetch} for physical experiments. 

\setlength{\tabcolsep}{5pt}
\begin{table}[tb]
    \caption{P2P Reward Components and Their AutoRL-Tuned Values }
	\label{tab:p2p_parameters}
    \centering
    \begin{tabular}{c|l|r|r|r|r}
        \hline \hline
        \textbf{Reward} & \multicolumn{1}{c|}{\textbf{Description}} & \multicolumn{1}{c|}{\textbf{Min}} & \multicolumn{1}{c|}{\textbf{Max}} & \multicolumn{1}{c|}{\textbf{Diff}} & \multicolumn{1}{c}{\textbf{Car}} \\
                        &                      &  &  & \multicolumn{1}{c|}{\textbf{Drive}} & \multicolumn{1}{c}{\textbf{Model}} \\ \hline
        \textbf{$r_\text{goal}$} & 1 when goal reached & 0.0 & 100.0 & \textbf{62.0} & \textbf{0.82} \\
         & and 0 otherwise &  &  &  \\
        \textbf{$r_\text{goalDist}$} & Negative Euclidean & 0.0 & 1.0 & \textbf{0.38} & N/A \\
         & distance to goal &  &  &  \\
        \textbf{$r_\text{goalProg}$} & Delta of Euclidean & 0.0 & 5.0 & N/A & \textbf{2.03} \\
         & distance to goal &  &  &  \\
        \textbf{$r_\text{collision}$} & 1 on collision & -100.0 & 0.0 & \textbf{-57.90} & \textbf{-1.80} \\
         & and 0 otherwise &  &  &  \\
        \textbf{$r_\text{clearance}$} & Distance to  & 0.0 & 1.0 & \textbf{0.67} & N/A \\
         & closest obstacle &  &  &  \\
        \textbf{$r_\text{step}$} & Constant per-step & -1.0 & 0.0 & \textbf{-0.43} & \textbf{-0.10} \\
         & penalty &  &  &  \\
        \textbf{$r_\text{turning}$} & Negative angular& 0.0 & 1.0 & \textbf{0.415} & N/A \\
                                    & speed & &  &  &  \\
        \textbf{$r_\text{backward}$} & Negative backward & -1.0 & 0.0 & N/A & \textbf{-0.64} \\
         & speed &  &  & &  \\ \hline \hline
    \end{tabular}
\end{table}
\setlength{\tabcolsep}{6pt}

\begin{table}[tb]
    \caption{Environments}
    \label{tab:environments}
    \centering
    \begin{tabular}{c|c|c|c}
    \hline \hline
    Environment & Type & Dimensions & Visual \\ \hline
    Training            &          Floor map & \dist{23} by \dist{18} & Fig. \ref{fig:mtv1965}   \\
    Building 1          &          Floor map & \dist{183} by \dist{66} & Fig. \ref{fig:b40}   \\
    Building 2          &          Floor map & \dist{60} by \dist{47} & Fig. \ref{fig:1667}   \\
    Building 3          &          Floor map &  \dist{134} by \dist{93} & Fig. \ref{fig:sfo}   \\
    Building Complex    &          Floor map & \dist{288} by \dist{163} & Fig. \ref{fig:googleplex}   \\
    Physical Testbed 1  &          SLAM &  \dist{50} by \dist{68} & Fig. \ref{fig:slam-map}   \\
    Physical Testbed 2  &          SLAM &  \dist{203} by \dist{135} & N/A (private)   \\
    Physical Testbed 3  &          SLAM &  \dist{22} by \dist{33} & Fig. \ref{fig:TB3-dense}   \\
    \hline \hline
    \end{tabular}
\end{table}

\begin{table*}[tb]
\caption{Baselines}
    \label{tab:baselines}
    \centering
    \begin{tabular}{c|c|c|c|c|c|l}
    \hline \hline
    Label   & Local    & Execution & \multicolumn{2}{c|}{Obstacle Avoid} & Monte Carlo & Description  \\
            & Planner (LP)  &  Policy   & LP & Execution & rollouts &              \\ \hline
    AutoRL  & N/A      &  AutoRL   & N/A & Yes & No & Local AutoRL policy like \cite{autorl} w/o guidance by a PRM. \\
    PRM-SL  & Straight Line & Straight  Line & No & No & No & Straight-line PRMs \cite{kavraki-prm} w/ straight-line execution policy.  \\
    PRM-GAPF  & Straight Line & Guided APF & No & Yes & No & Straight-line PRMs \cite{kavraki-prm} executed by guided APF like \cite{chiang2015path}.\\
    PRM-DWA  & Straight Line & Guided DWA & No & Yes & No & Straight-line PRMs \cite{kavraki-prm} executed by guided DWA \cite{fox1997dynamic}.\\
    \textbf{PRM-RL}  & \textbf{AutoRL}   & \textbf{AutoRL}   & \textbf{Yes} & \textbf{Yes} & \textbf{Yes} & \textbf{PRM-RL w/ AutoRL for roadmap \& execution (ours).} \\ \hline
    PRM-HTRL  & DDPG     &  DDPG     & Yes & Yes & Yes & PRM-RL w/ DDPG for roadmap \& execution (\cite{prm-rl}). \\
    SF  & N/A      &  SF   & N/A & Yes & No & Successor Features (SF) visual local planner \cite{successor-features}. \\
    \hline \hline
    \end{tabular}
\end{table*}

\subsubsection{Obstacle-avoidance local planner training}
We train P2P agents with AutoRL over DDPG \cite{autorl} with reward and network tuning for the differential drive robot, and reward tuning for the car robot. In both cases, the true objective for training the local planner is to navigate within \dist{0.25} of the goal, allowing the RL agent to cope with noisy sensors and dynamics. Table \ref{tab:p2p_parameters} depicts learned reward hyperparameters. DDPG actor and critic are feed-forward fully-connected networks. Actor networks are three layers deep, while the critics consists of a one or two-layer observation networks joined with the action networks by two fully connected layers. Actor, critic joint, and critic observation layer widths are $(241, 12, 20) \times (607, 242) \times (84)$ for the reward-and-network trained differential drive model and $(50, 20, 10) \times (10 ,10) \times (50, 20)$ for the reward-trained car model. Appendix \ref{app:training} contains the training hyperparameter details.
The training environment is \dist{14} by \dist{17} (Fig. \ref{fig:mtv1965}). To simulate imperfect real-world sensing, the simulator adds Gaussian noise, $\gauss{0}{0.1}$, to its observations.


\subsubsection{Evaluation environments}
\begin{figure*}[tb]
	\begin{center}
		\begin{tabular}{ccc}
            \subfloat[SLAM Map---\dist{50} by \dist{68}]{\includegraphics[trim=0mm 0mm 0mm 0mm,clip,width=5.5cm,keepaspectratio=true]{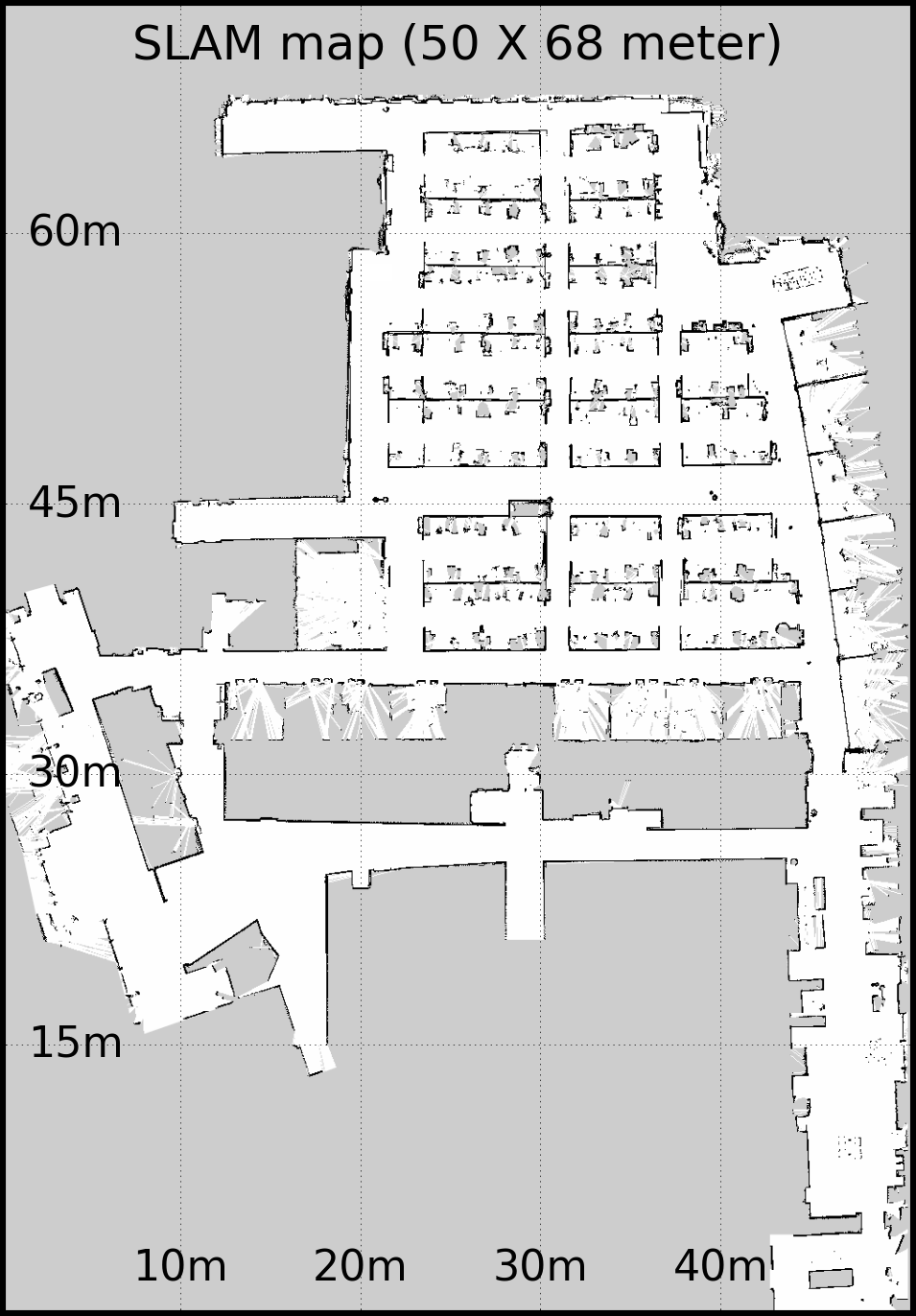}\label{fig:slam-map}} 
			&
            \subfloat[Roadmap---0.4 samp. / \dist{} at $90\%$ succ.]{\includegraphics[trim=0mm 0mm 0mm 0mm,clip,width=5.5cm,keepaspectratio=true]{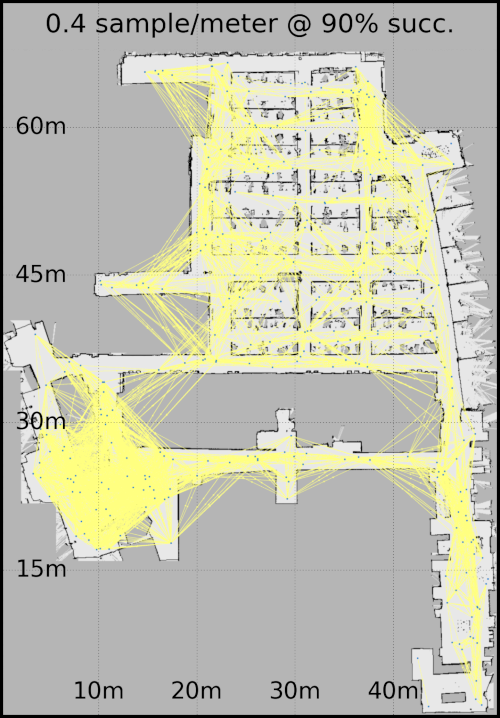}\label{fig:slam_roadmap}}
            &
            \subfloat[Robot Trajectories---12 with $83\%$ success]{\includegraphics[trim=0mm 0mm 0mm 0mm,clip,width=5.5cm,keepaspectratio=true]{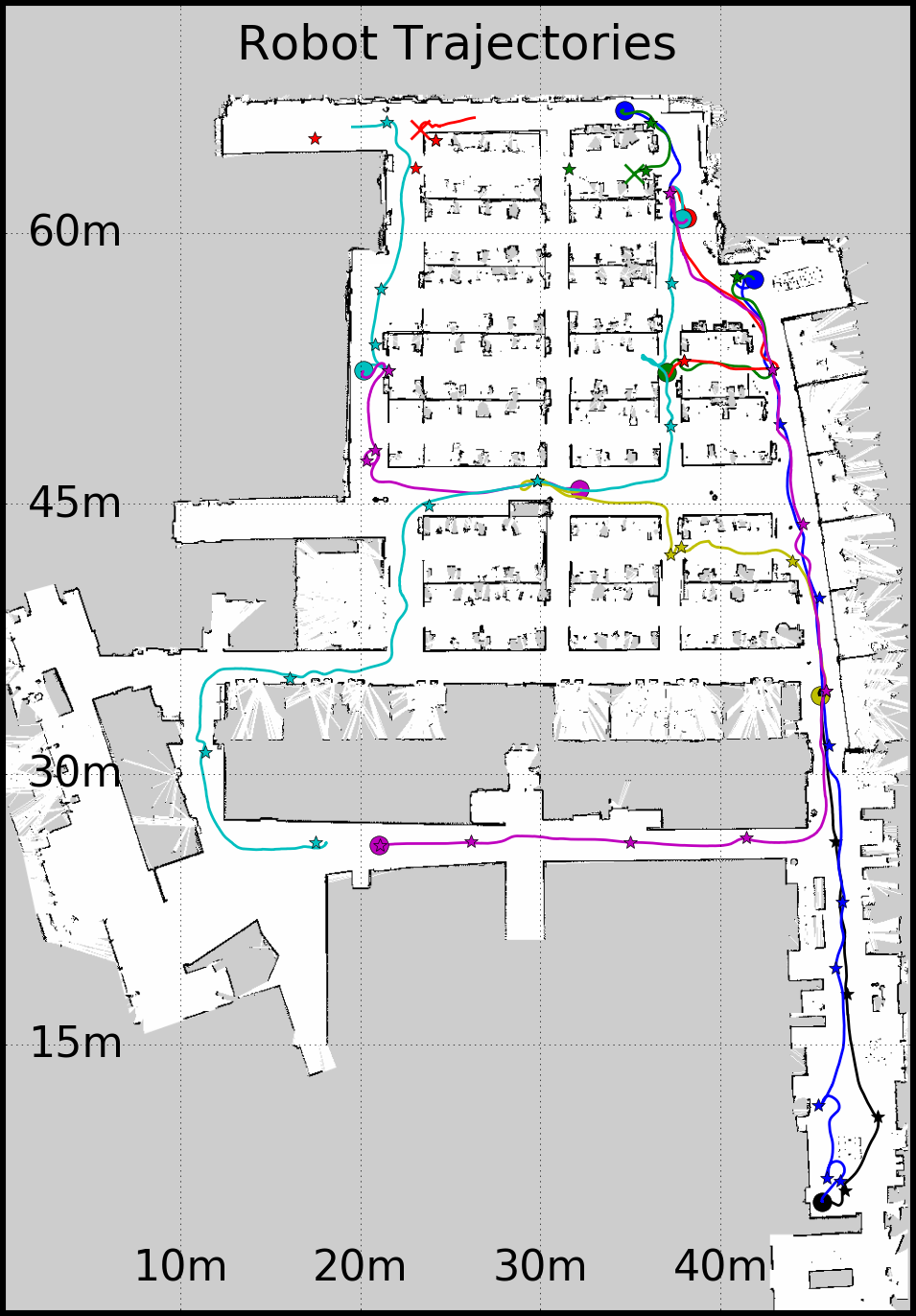}\label{fig:slam_trajectories}}\\
		
	    \end{tabular}
		\caption{\small PRMs for robot deployments are built over SLAM maps of the target environment. (a) SLAM map collected at the site of a robot deployment. (b) PRM-RL learns the effective connectivity of the map for the policy. (c) PRM-RL achieves $83\%$ success on the robot.\label{fig:slam}}
	\end{center}
\end{figure*}

Table \ref{tab:environments} lists our simulation environments, all derived from real-world buildings. \textit{Training}, \textit{Building 1-3}, depicted in Fig. \ref{fig:maps}, and \textit{Building Complex} (Fig. \ref{fig:googleplex}) are metric maps derived from real building floor plans. They are between 12 to 200 times larger than the training environment by area. \textit{Physical Testbed 1} (Fig. \ref{fig:slam}), \textit{Physical Testbed 2} and \textit{Physical Testbed 3} (Fig. \ref{fig:TB3-dense}) are derived from SLAM maps used for robot deployment environments. 

\subsubsection{Roadmap building}
For simplicity we use uniform random sampling to build roadmaps. We connect PRMs with a $p_{s}$ effective threshold of $\geq 90\%$ over 20 attempts, with a max connection distance $d_{\omega}$ of \dist{10} based on the effective navigation radius $d_{\pi}$ for our P2P agents per \cite{autorl}, except where otherwise stated.

\subsubsection{Baselines}
\label{sec:baselines}
Table \ref{tab:baselines} shows four selected baselines. The baselines differ in the local planner, used for building the roadmap, and the execution policy, which guides the robot. We select baselines given their ability to avoid obstacles and deal with stochasticity. Recall that PRM-RL relies on a stochastic policy capable of obstacle-avoidance to connect nodes in the roadmap using  Monte Carlo rollouts of the policy.

The baselines for experimental comparison include a local planner based on AutoRL \cite{autorl}, PRM-SL \cite{kavraki-prm}, PRM-GAPF (a modification of \cite{chiang2015path}), and PRM-DWA. PRM-SL \cite{kavraki-prm} uses roadmaps built with a straight-line planner and a straight-line execution policy. PRM-GAPF uses PRMs built with a straight-line planner, and an execution policy of APF, an artificial potential field planner \cite{khatib1986real}, guided by the PRM-SL path, similar to \cite{chiang2015path}. PRM-DWA is similar to PRM-GAPF using DWA, a dynamic window avoidance local planner \cite{fox1997dynamic}.

In addition, we compare against two other baselines numerically: PRM-HTRL \cite{prm-rl} is our original PRM-RL with hand-tuned DDPG as the planner; where not otherwise specified, PRM-RL refers to our current approach. Successor Features \cite{successor-features} is a visual-based navigation approach using discretized actions. We do not compare PRM-RL with RRTs here because this work focuses on solving the multi-query problem, while RRTs solve single-query problems, making them comparatively expensive for building long-range trajectories on the fly, especially when incorporating end-to-end controls. While large roadmaps can be expensive to construct, they can be reused for many queries across many robots, and queries are fast. For example, RRT solves a single query in the same environment in about 100 seconds \cite{rlrrt}, while PRM-RL produces a path in less than a second for a pre-built roadmap (see Table \ref{tab:prm-rl-vs-baselines}), although the roadmap takes hours to build.

For comparisons to baselines, unless otherwise stated, each roadmap is evaluated on 250 queries selected from the $\cfree$ between start and goal positions from \dist{1.5} to \dist{100}. We measure start to goal distance by the shortest feasible path as estimated by our simulation framework using a queen's-move discretized A* search with adequate clearance for the robot to travel without collision.

\begin{table*}[tb]
\caption{PRM-RL Performance vs Baselines.}
    \label{tab:prm-rl-vs-baselines}
    \centering
    \scriptsize
    \begin{tabular}{cc|rrr|rr|rr|rr|rr|rr|r}
    \hline \hline
 Baseline & PRM &  Succ. & 99\% & \multicolumn{1}{r|}{PRM-RL} & \multicolumn{2}{c|}{Path Dist. (m)} & \multicolumn{2}{c|}{Path / Shortest.} & \multicolumn{2}{c|}{Clearance (m)} & \multicolumn{2}{c|}{Time (s)} &  \multicolumn{2}{c|}{Roadmap} &\multicolumn{1}{c}{Coll Checks} \\
  & Density &  \% &  Conf. &  \multicolumn{1}{r|}{Nav$\Delta$\%} & Mean &  Std. &  Mean &  Std. &  Mean &  Std. &  Plan. &  Exec.  & Nodes & Edges & (x$10^6$)  \\     \hline
   AutoRL &          N/A &         8.53 &               1.52 &              83.20 &                  \best{14.2} &                        \best{15.9} &              1.98 &                    5.98 &              0.43 &                    0.57 &          \best{0.03} &           42.4 & N/A & N/A &              N/A \\ \hline
   PRM-SL &       Sparse &        12.40 &               1.79 &              79.33 &                  28.6 &                        26.2 &              1.16 &                    0.63 &              0.28 &                    0.14 &          0.07 &           16.7  & \best{1321} & 28984 &          \best{1.4} \\
   PRM-SL &        Dense &         9.47 &               1.59 &              82.27 &                  20.3 &                        17.3 &              \best{1.10} &                    \best{0.48} &              0.28 &                    0.13 &          0.43 &           \best{12.8}  & {3302} & {186491} &          9.0 \\ \hline
 PRM-GAPF &       Sparse &        29.20 &               2.47 &              62.53 &                  57.2 &                        42.5 &              1.35 &                    0.70 &              0.30 &                    0.13 &          0.06 &           44.6 & {\best{1321}} & {28984} &          \best{1.4} \\
 PRM-GAPF &        Dense &        28.13 &               2.45 &              63.60 &                  47.0 &                        35.9 &              1.36 &                    0.65 &              0.30 &                    0.13 &          0.39 &           38.1 & {3302} & {186491} &          9.0 \\ \hline
  PRM-DWA &       Sparse &        40.53 &               2.67 &              51.20 &                 277.6 &                       192.6 &              7.23 &                   17.09 &              \best{0.48} &                    0.11 &          0.06 &          167.1 & {\best{1321}} & {28984} &          \best{1.4} \\
  PRM-DWA &        Dense &        40.13 &               2.67 &              51.60 &                 266.2 &                       215.7 &              6.44 &                    8.20 &              0.47 &                    \best{0.10} &          0.40 &          178.8 & {3302} & {186491} &          9.0 \\ \hline
   PRM-RL &       Sparse &        86.53 &               1.86 &               5.20 &                  61.5 &                        36.7 &              1.16 &                    1.90 &              0.43 &                    0.18 &          0.14 &           70.4 & {\best{1321}} & {55270} &          74.0 \\
   PRM-RL &        Dense &        \best{91.73} &               \best{1.50} &                \best{Best} &                  60.0 &                        36.5 &              \best{1.10} &                    1.82 &              0.42 &                    0.14 &          0.75 &           62.7 & {3302} & {\best{332399}} &          360.0 \\
    \hline \hline
    \end{tabular}
\end{table*}

\begin{figure*}[t]
	\begin{center}
		\begin{tabular}{cc}
			\subfloat[Success on floorplan maps]{\includegraphics[width=0.48\textwidth]{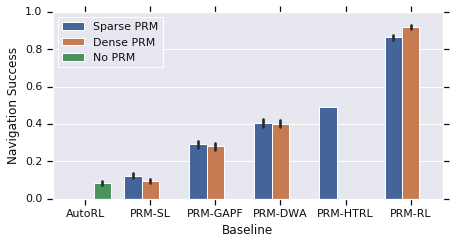}\label{fig:sxs-floorplan}}
			&
			\subfloat[Success on SLAM maps]{\includegraphics[width=0.48\textwidth]{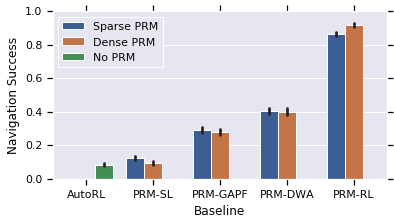}\label{fig:sxs-slam}}\\
		\end{tabular}
		\caption{\small PRM-RL outperforms the AutoRL local planner as well as PRM-SL, PRM-guided versions of GAPF and DWA, and our prior work PRM-HTRL. (a) PRM-RL's success is up to $93\%$ on floorplan maps. (b) PRM-RL's success is up to $89\%$ on SLAM maps.}
	\end{center}
\end{figure*}

\begin{figure*}[t]
	\begin{center}
		\begin{tabular}{c}
        \includegraphics[trim=0mm 0mm 0mm 0mm,clip,width=1.0\textwidth,keepaspectratio=true]{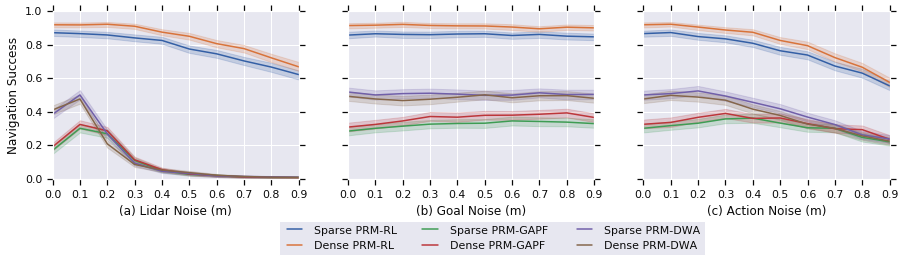}
		\end{tabular}
		\caption{\small PRM-RL is more robust to noise than PRM-GAPF or PRM-DWA. (a) As lidar noise increases, PRM-RL degrades slowly, showing a $28\%$ drop at a noise level of \dist{1}, whereas PRM-GAPF and PRM-DWA degrade quickly to roughly $1\%$ performance at noise of \dist{0.75}. (b) All methods show resistance to position noise (modeled as goal uncertainty). (c) As action noise increases, PRM-RL degrades slowly, showing a $37\%$ drop at noise of \dist{1}; PRM-GAPF and PRM-DWA are degraded up to $39\%$ and $54\%$ of their peak performance, respectively. \label{fig:robustness-to-noise}}
	\end{center}
\end{figure*}

\begin{figure*}[t]
	\begin{center}
		\begin{tabular}{cc}
		\subfloat[Success rate over sampling density]{\includegraphics[width=0.48\textwidth]{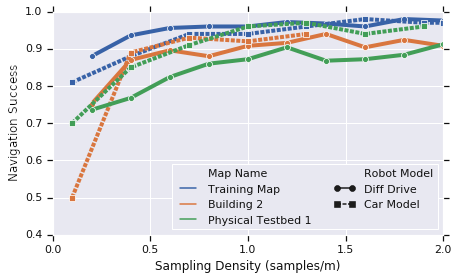}\label{fig:dens-vs-eval}}
			&
		\subfloat[Collision Checks over sampling density]{\includegraphics[width=0.48\textwidth]{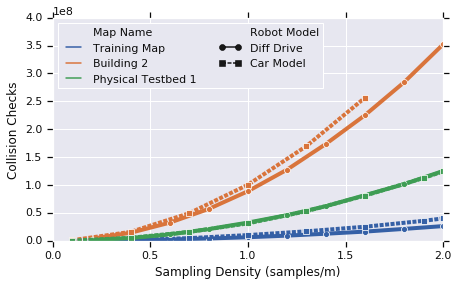}\label{fig:dens-vs-checks}}\\
		\end{tabular}
		\caption{\small Increasing sampling density improves performance at the cost of roadmap construction time. (a) As density increases, RL agents guided by PRM-RL succeed more often with a sweet spot of 1 node per square meter. (b) Cost rises prohibitively as sampling rises; over 1 node per square meter, collision checks for the training map surpass our largest floorplan roadmap collected at 0.4 per square meter. \label{fig:sampling-density}}
	\end{center}
\end{figure*}

\begin{figure*}[t]
	\begin{center}
		\begin{tabular}{c}
        \includegraphics[trim=0mm 0mm 0mm 0mm,clip,width=0.95\textwidth,keepaspectratio=true]{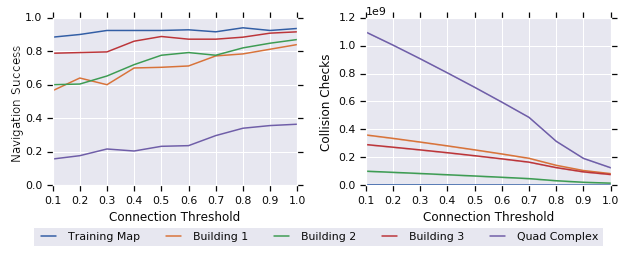}
		\end{tabular}
		\caption{\small Increasing required edge connection success improves performance and reduces collision checks. (a) As the threshold for connecting nodes in the PRM rises, RL agents guided by PRM-RL succeed more often with a sweet spot of $90\%$ and higher. (b) Furthermore, early termination enables PRM-RL to skip unneeded connectivity checks for savings exceeding $60\%$, an effect important on large roadmaps. \label{fig:success-sweeps}}
	\end{center}
\end{figure*}

\subsection{PRM-RL Performance on Floorplan Maps}
\label{sec:results-floorplan}

Table \ref{tab:prm-rl-vs-baselines} shows PRM-RL's performance compared to baselines on Buildings 1-3 of our floorplan maps, using both sparse and dense PRMs; Fig.  \ref{fig:sxs-floorplan} also shows for reference our prior work PRM-HTRL. PRM-RL's average success rate in the dense condition is $91.7\%$ over all three maps, which outperforms the baselines by  $83.2\%$ for pure AutoRL, $82.2\%$ for dense PRM-SL, and $51.6\%$ for dense PRM-DWA. Outperforming AutoRL's non-guided local policies is not surprising as they do not have the knowledge of the obstacle map, but we include it for completeness. Successful paths executed by AutoRL are shortest, but dense PRM-SL and dense PRM-RL are within $10\%$ of the shortest feasible path as estimated by our simulation framework. DWA path lengths are much longer because DWA exhibits safe looping behavior in box canyons given the action space and noisy observations, but DWA maintains the best clearance to obstacles. 

Analyzing collision checks (the complexity of building a roadmap), planning time, and execution time reveals interesting tradeoffs. First, PRM-RL requires 1-2 orders of magnitude more collision checks than PRM-SL; increasing map density only consistently helpes PRM-RL. This is because of the Monte Carlo rollouts. At runtime, path finidng requires no collision checking, as it performs a graph search on a pre-built roadmap. At runtime, planning with a roadmap requires no collision checks, though local control policies may carry out collision checks in execution (e.g., DWA collision checks trajectories in velocity space). Second, PRM-RL planning time is up to twice as long as PRM-SL, because the PRM-RL roadmaps contain more edges due to the RL local planner connecting to nodes that are not in the clear line of sight. Still, planning takes less than a second even for the densest maps, an insignificant part of execution time in both simulation and  robot deployment. Last, the execution time for PRM-RL is almost five times longer than PRM-SL because PRM-RL succeeds at longer trajectories thanks to RL control policy that adapts on-the-go to uncertainties not present in planning time (i.e. moving obstacles or sensor noise).

We can draw three observations from these results. First, guiding a local planner with PRM-RL can vastly improve the planner's success rate; this is not surprising as our local planners are not designed to travel long distances. Second, PRM-RL successfully enables local planners to transfer to new environments: the AutoRL policy only saw the Training Map in training, yet performs at a $91\%$ success rate on our evaluation maps.  Third, the PRM success rate is correlated with the success of the local planner. PRM-RL with an AutoRL policy and a sparse PRM build achieves $86.5\%$ success, a $37.5\%$ increase over the sparse $49\%$ success rate reported in PRM-HTRL. This is evidence that investing in a high-fidelity local planner increases PRM-RL's performance of the overall navigation task. In contrast, moving to a denser PRM map (which we discuss in more detail in Section \ref{sec:results-connect}) provides a lesser increase of $5.2\%$; this is still significant for deployment, however, as it also represents a $38.6\%$ decrease in errors.

\subsection{PRM-RL Robustness to Noise}
\label{sec:results-noise}
Sensors and actuators are not perfect, so navigation methods should be resilient to noise. Fig. \ref{fig:robustness-to-noise} shows the evaluation of PRM-RL, PRM-GAPF, and PRM-DWA on the Training map, Buildings 1-3 and Physical Testbed 1 over simulated Gaussian noise sources with mean $0.0$ and $\sigma$ in the following ranges:

\begin{itemize}
    \item \textbf{Lidar sensor noise} $\sigma_l$ from \dist{0}-\dist{0.9}, over three times the radius of the robot.
    \item \textbf{Goal position noise} $\sigma_g$ from \dist{0}-\dist{0.9}, over three times the radius of the goal target.
    \item \textbf{Action noise} of velocity $\sigma_v$ \velocity{0}-\velocity{0.9} and angular velocity $\sigma_a$ \angvel{0.9}.
\end{itemize}

As lidar and action position noise increase, PRM-RL shows only a slight degradation of $28\%$ on lidar noise and $37\%$ on action noise, even at \dist{0.9}.  In contrast, PRM-GAPF and PRM-DWA degrade steeply with respect to lidar noise, with success rates dropping to less than $1\%$. These methods seem to be more resistant to increased action noise, but still drop to $39\%$ and $54\%$ of their peak scores for PRM-GAPF and PRM-DWA respectively. All methods were relatively resistant to goal noise, with less than $10\%$ falloff. PRM-RL outperformed PRM-GAPF and PRM-DWA in all conditions.

PRM-RL is resilient to lidar, localization, and actuator noise on the order of tens of centimeters, which is larger than the typical lidar, localization, and actuator errors we observe on our robot platforms, indicating that PRM-RL is a feasible approach to deal with the kinds of errors our robots actually encounter. This is similar to the trend to noise sensitivity reported in \cite{autorl}, and suggests that overall navigation resilience to noise is correlated to that of the local planner.

\subsection{The Impact of Sampling Density and Success Threshholds}
\label{sec:results-connect}
To deploy PRM-RL on real robots, we need to choose sampling density and success threshold parameters that provide the best performance. Fig.  \ref{fig:sampling-density} shows that PRM-RL success rate increases steadily up to a sampling density of $1.0$ samples per meter, which is roughly twice the size of our robot, and then levels off. At the same time, collision checks increase rapidly with sampling density; we have observed that over $1.0$ samples per meter, the roadmap for the training environment requires more collision checks than some of the larger evaluation roadmaps collected at $0.4$ samples per meter. While PRM-SL theory predicts that performance would continue to improve with increased sampling \cite{Kavraki98probabilisticroadmaps}, \cite{hsu2007probabilistic}, this suggests that beyond a critical density PRM-RL performance is robot-noise-bound, and that sampling beyond this density provides little benefit at rapidly increasing cost.

These experiments evaluate global planning; however, note the average distance of points at 2.0 samples per square meter is roughly \dist{0.5}, the true objective distance used to train our AutoRL local planners.  This suggests the asymptotic behavior in these experiments could be explained by the PRM sampling density approaching a distance where the local planner can almost always find a path to a nearby PRM node. Also note the car model's success rate differs from the diff-drive model, suggesting that success rate is model bound rather than map bound.

Fig.  \ref{fig:success-sweeps} shows that PRM-RL's success rate increases with the required connection success rate. Because our connection algorithm terminates earlier for higher thresholds when it detects failures, collision checks drop as the success threshold rises. At the end, for larger roadmaps, the success threshold of 100\% not only produces the most reliable roadmaps, but requires fewer collision checks to build them.

These results suggest choosing map densities up to $1.0$ samples per meter with as high a success connection threshold as possible. In this paper, we compare sparse and dense parameters: a sampling density of $0.4$ samples per meter and an effective success connection threshold of $\geq 90\%$  which enables comparison with \cite{prm-rl}, and a density $1.0$ samples per meter with a threshold of $100\%$; in almost all evaluations, we observe better performance with the dense parameters.

\subsection{PRM-RL Performance on SLAM Maps}
\label{sec:results-slam}
Floorplans are not always available or up-to-date, but many robots can readily generate SLAM maps and update them as buildings change. To assess the feasibility of using SLAM maps, we evaluate PRM-RL on SLAM maps generated from some of the same buildings as our floorplan roadmaps. Fig.  \ref{fig:slam} illustrates a sample roadmap built from a SLAM map generated with the ROS distribution of the GMapping algorithm \cite{gmapping} with the default resolution of 5cm per pixel. This map corresponds to \textit{Physical Testbed 1}, part of floor two of Building 1 in Fig. \ref{fig:b40} and the upper center section of the large-scale roadmap in Fig.  \ref{fig:googleplex}. This SLAM-derived roadmap has 195 nodes and 1488 edges with 2.1 million collision checks.

We compare PRM-RL with our baselines on the three Physical Testbed SLAM maps. PRM-RL's success rate is $89\%$ on the dense PRM, a $97\%$ relative increase over PRM-DWA and a $157\%$ increase over PRM-GAPF. 

These results show that the performance of PRM-RL with an AutoRL policy is lesser but comparable to its performance on floorplan maps, and exceeds all other baselines; it is even superior to PRM-HTRL on floorplan maps. These results indicate PRM-RL performs well enough to merit tests on roadmaps intended for real robots at physical sites, which we discuss in the following two sections.

\subsection{Scaling PRM-RL to Large-Scale Maps}
\label{sec:results-large}
Our robot deployment sites are substantially larger than our simulated test maps, raising the question of how PRM-RL would scale up. For example, the SLAM map discussed in the previous section is only part of one building within a quad-building complex. Where the SLAM map is \dist{78} by \dist{44}, a map of the quad-building complex is \dist{288} by \dist{163}. To assess PRM-RL's performance on large-scale maps, we build and test roadmaps for maps covering all deployment sites.

Fig.  \ref{fig:googleplex} depicts a large floorplan roadmap from the quad-building complex. This roadmap has 15,900 samples and 1.4 million candidate edges prior to connection attempts, of which 689,000 edges were confirmed at a $90\%$ success threshold. This roadmap took 4 days to build using 300 workers in a cluster, and required 1.1 billion collision checks. PRM-RL successfully navigates this roadmap $57.3\%$ of the time, evaluated over 1,000 random navigation attempts with a maximum path distance of \dist{1000}. Note our other experiments use a maximum path distance of \dist{100}, which generally will not cross the skybridge in this larger map. For reference, using our default evaluation settings, PRM-RL navigates this roadmap $82.3\%$ of the time.

For our other candidate robot deployment site, we use a large SLAM map, \dist{203} by \dist{135}. We constructed a roadmap with 2069 nodes and 53,800 edges, collected with 42 million collision checks at the higher success threshold of $100\%$. PRM-RL successfully navigated this $58.8\%$ of the time, evaluated over 1,000 random navigation attempts. As on our smaller SLAM map, the failure cases indicate that the more complex geometry recorded by SLAM proves problematic for our current policies.

These results indicate that PRM-RL's simulated performance on large-scale roadmaps surpasses the average success threshold we observed previously in \cite{prm-rl}, making it worthwhile to test on real robots at the physical sites.

\begin{figure*}[t]
	\begin{center}
		\begin{tabular}{ccc}
            \subfloat[Physical Testbed 2, Dense  Map]{\includegraphics[trim=0mm 0mm 0mm 0mm,clip,height=4.25cm,keepaspectratio=true]{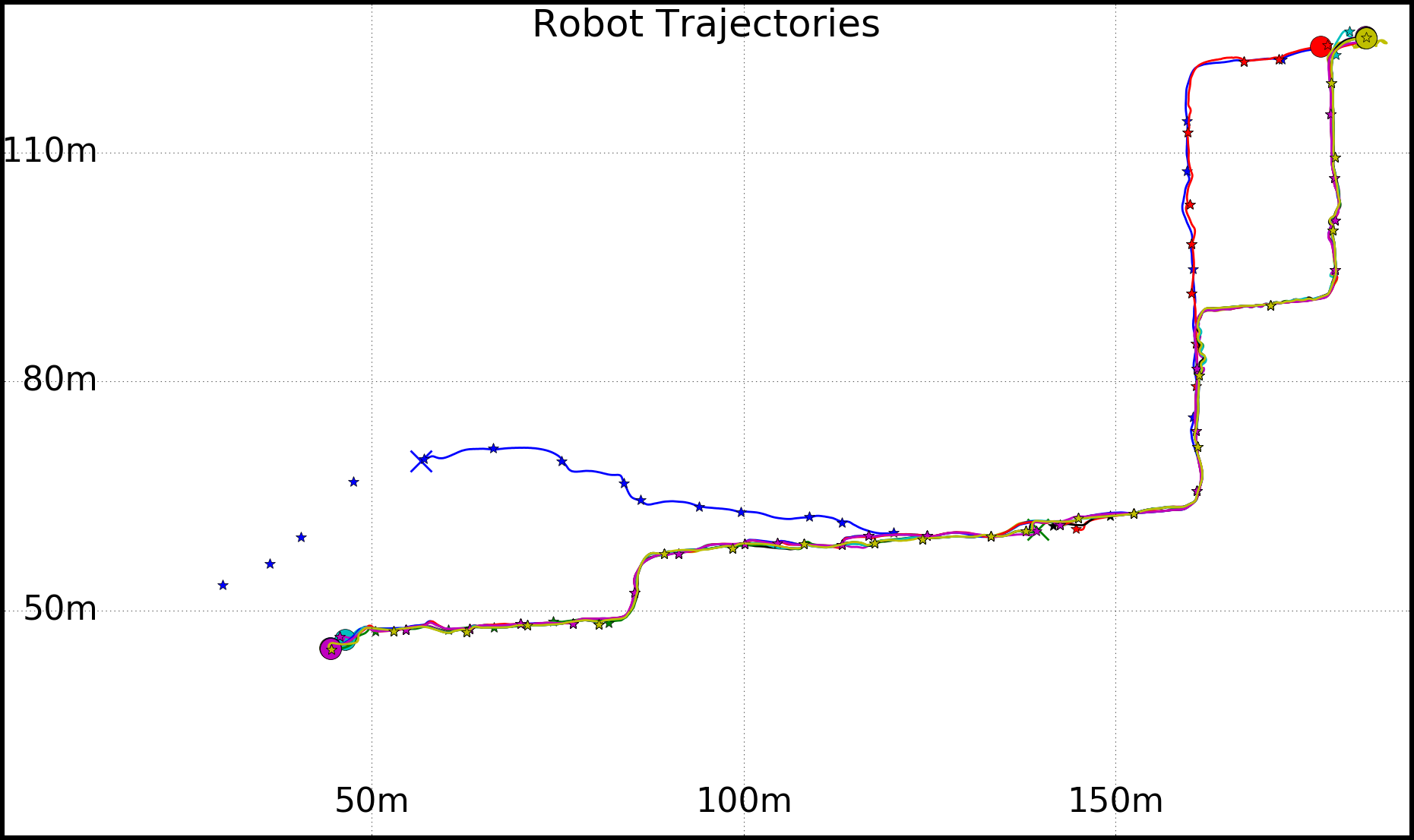}\label{fig:TB2-dense}}
			&
            \subfloat[Physical Testbed 3, Sparse  Map]{\includegraphics[trim=0mm 0mm 0mm 0mm,clip,height=4.25cm,keepaspectratio=true]{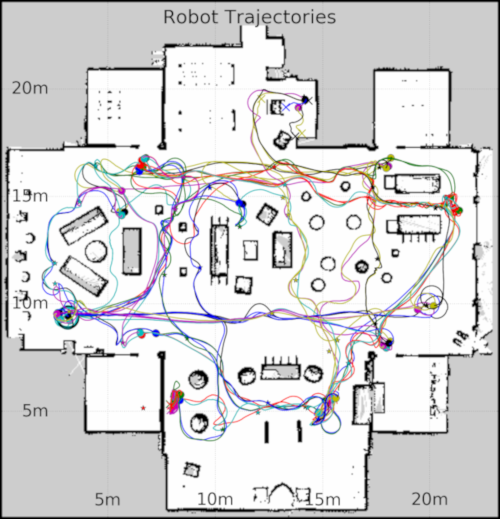}\label{fig:TB3-sparse}}
			&
            \subfloat[Physical Testbed 3, Dense  Map]{\includegraphics[trim=0mm 0mm 0mm 0mm,clip,height=4.25cm,keepaspectratio=true]{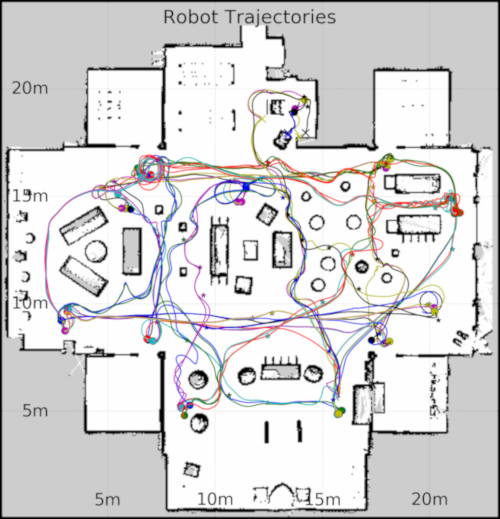}\label{fig:TB3-dense}}\\
		\end{tabular}
		\caption{\small Trajectories collected from PRM-RL execution on real robots. Trajectories are in color, circles represent successful navigation, X's represent emergency stops. (a) Several queries executed on a differential drive robot and tracked with onboard localization in a real office environment in \textit{Physical Testbed 2}.  The longest successful single trajectory was \dist{221.3}. The floorplan and PRM connectivity are not displayed for privacy.  (b)-(c) \textit{Physical Testbed 3} is a simulated living room, where 128 trajectories collected with sparse and dense map densities. The denser map achieved $6\%$ higher success rate and produced on average \dist{2.5} shorter paths.  \label{fig:real-robot}}
	\end{center}
\end{figure*}

\begin{figure}[tb]
    \centering
    \includegraphics[width=0.48\textwidth,keepaspectratio=true]{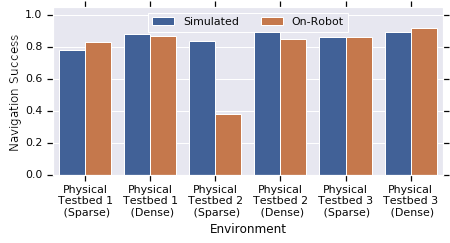}
    \caption{\small PRM-RL closes the sim to real gap. Performance in simulation and on robots is similar, despite an e-stop policy that stops robots more aggressively than in simulation. Note that on Physical Testbed 2, a dense PRM overcomes an obstacle that thwarted a sparse PRM.\label{fig:sim-to-real}}
\end{figure}

\begin{figure*}[t]
	\begin{center}
		\begin{tabular}{cc}
	        \subfloat[Roadmap built using car model on Training (\dist{23} by \dist{18})]{\includegraphics[trim=0mm 0mm 0mm 0mm,clip,height=6.75cm,keepaspectratio=true]{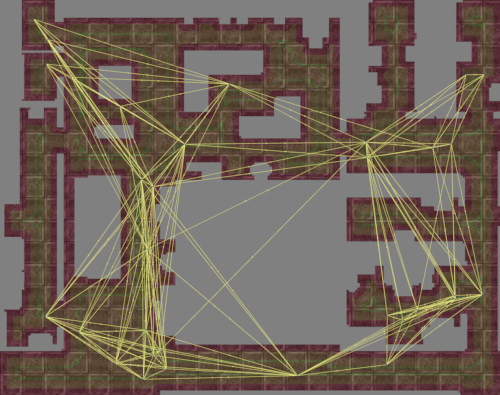}\label{fig:car-roadmap}} 
		    &
			\subfloat[Example car model trajectory on Training (\dist{23} by \dist{18})]{\includegraphics[trim=0mm 0mm 0mm 0mm,clip,height=6.75cm,keepaspectratio=true]{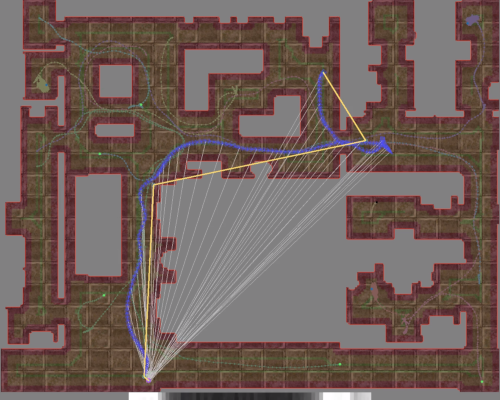}\label{fig:car-trajectory}}\\
		\end{tabular}
		\caption{\small PRMs can be built for agents with dynamic constraints. (a) Roadmap built for our training environment using a nonholonomic car model. (b) Example car model trajectory; the upper right shows a three point turn to change the robot orientation. Yellow lines are the PRM path, the blue line is the agent's trajectory, white lines indicate progress towards the goal, and light green dots represents previous evals. \label{fig:car-model}}
	\end{center}
\end{figure*}

\begin{figure}[tb]
    \centering
    \includegraphics[width=0.48\textwidth,keepaspectratio=true]{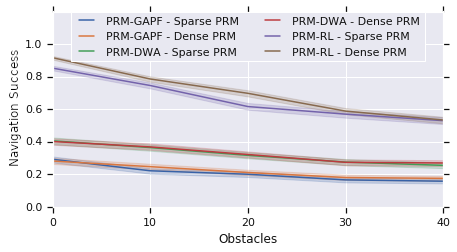}
    \caption{\small PRM-RL is resilient in the presence of dynamic obstacles. With $40$ moving agents, performance degrades to $53.1\%$, a higher score than PRM-GAPF and PRM-DWA achieve in the default condition. \label{fig:dynamic-obstacles}}
\end{figure}

\begin{figure}[tb]
    \centering
    \includegraphics[width=0.48\textwidth,keepaspectratio=true]{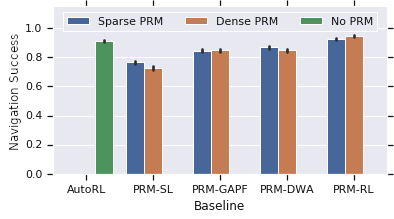}
    \caption{\small PRM-RL is competitive with Successor Features \cite{successor-features}. On transfer maps, PRM-RL achieves over $94\%$ success with noisy lidar, continuous actions, and arbitrary goals, whereas Successor Features reports $98\%$ success with vision, discrete action and a single goal. \label{fig:sxs-synthetic}}
\end{figure}

\subsection{Transfer of PRM-RL to Physical Robots}
\label{sec:results-robots}
We empirically evaluate PRM-RL on three physical environments on two differential-drive robots. First, we evaluate PRM-RL with the AutoRL policy for a differential-drive robot in \textit{Physical Testbed 1} (Fig. \ref{fig:slam-map}). 
We collected 27 trajectories over \dist{831} of travel with an overall success rate of $85.2\%$; the longest successful trajectory was \dist{83.1} meters. Fig. \ref{fig:slam_trajectories} shows the trajectories of 12 runs.

Second, we evaluate PRM-RL in \textit{Physical Testbed 2}. For a variant of the roadmap generated at $90\%$ success rate with a density of $0.4$ samples per meter, we collected 8 trajectories over \dist{487.2} of travel; the longest successful trajectory was \dist{96.9}. We cannot directly compare this evaluation to our simulated runs because the e-stop policies designed to protect the robot do not match our simulation termination conditions. Nevertheless, we recorded 3 successful runs out of 8, a $37.5\%$ success rate. We then tested a variant of the roadmap generated at $100\%$ success rate and a density of $1.0$ samples per meter over 13 runs on  \dist{2542.1} of travel for an improved $84.6\%$ success rate, shown in Fig. \ref{fig:TB2-dense}; the longest successful trajectory was \dist{221.3}.

Third, we apply the same evaluation parameters from \textit{Physical Testbed 2} onto a novel environment, \textit{Physical Testbed 3}. We collected 128 runs, 64 for each variant of the roadmap, testing both with the same sets of start/goal points. The denser map achieved a $92.2\%$ success rate compared to $85.9\%$ for the sparse map. Since \textit{Physical Testbed 3} contains large sections of free space, using a denser map resulted in more optimized node connections and reduced robot traversal by an average of \dist{2.5} per run.

Fig.  \ref{fig:sim-to-real} summarizes these results. Despite more aggressive episode termination policies on robots (near-collisions are treated as failures), we nonetheless observe similar results: over several different roadmaps constructed at different densities and success criteria, PRM-RL achieves $85.8\%$ success on robots with an average sim2real gap of $7.43\%$. These results show that the performance of PRM-RL on robots is correlated with the performance of PRM-RL in simulation. This makes PRM-RL a useful method for closing the sim-to-real gap, as improvements or regressions in simulation are likely to be reflected in performance on robots.

\subsection{PRM-RL with Kinodynamic Constraints}
\label{sec:results-car}
To demonstrate that PRM-RL could guide more complicated robots, we develop a drive model for a simulated F1/10 car \cite{f1tenth} with dynamics following \cite{DBLP:journals/corr/PadenCYYF16} with velocity and acceleration constraints enforced by our simulator. In this setup, AutoRL learns a steering function for a robot with kinodynamical constrains, effectively learning to respect velocity and acceleration constraints. Algorithm \ref{alg:add_edge} encodes those constrains globally by connecting only reachable nodes, making PRM-RL effectively a kinodynamic planner \cite{Donald:1993:KMP:174147.174150}. Average success over the four maps in Fig.  \ref{fig:maps} is $85.8\%$ with a standard deviation of $1.0\%$; average success in simulation on \textit{Physical Testbed 1} with a goal distance of \dist{0.25} is $85.8\%$. Fig.  \ref{fig:car-roadmap} illustrates a roadmap built with this model over our training map with $0.4$ samples per meter connected with a $90\%$ success rate; this roadmap has 32 nodes and 313 edges connected after 403,000 edge checks. On this roadmap, PRM-RL exhibits an $83.4\%$ success rate, including cases where the car needs to make a complex maneuver to turn around (Fig. \ref{fig:car-trajectory}). These results are comparable to results on the robot, indicating that PRM-RL is a viable candidate for further testing on more complicated robot models.

\subsection{PRM-RL with Dynamic Obstacles}
\label{sec:results-dynamic}

PRM-RL is also resilient in the face of dynamic obstacles, relying on the local planner to avoid them without explicit replanning. We simulated pedestrians with the social force model \cite{helbing1995social} and tested PRM-RL, PRM-GAPF, and PRM-DWA on Buildings 1-3 with 0 to 40 added agents. While all methods showed a similar degradation in the presence of obstacles, on average $39.0\%$, PRM-RL's performance only dropped to $53.1\%$ with 40 obstacles, superior to PRM-GAPF ($28.7\%)$ and PRM-DWA ($40.3\%$) even in the zero-obstacle condition. Thus, a resilient local planner can enable PRM-RL to handle dynamic obstacles even though the framework has no explicit support for dynamic replanning.

\subsection{PRM-RL in Synthetic Environments}
\label{sec:results-synthetic}
The Successor Features visual navigation approach achieves $100\%$ success on training maps and $98\%$ success on transfer to new maps \cite{successor-features}. Successor Features navigates to single targets using vision and a discretized action space. In contrast, PRM-RL navigates to arbitrary targets using lidar and a continuous action space. Nevertheless, both approaches navigate in spaces qualitatively similar to each other in simulation and in testing on robots.

Therefore, we evaluate PRM-RL and our baselines in simulation on close approximations of the maps used in \cite{successor-features}, which include four maze-like simulated maps and two physical testbed environments. Fig. \ref{fig:sxs-synthetic} shows the results: Both AutoRL and PRM-RL achieved over $90\%$ success on the synthetic maps, with PRM-RL on a dense map achieving $94.6\%$ success. While Successor Features's $98\%$ success is $3.4\%$ higher than PRM-RL, it has access to a visual sensor, executes discretized actions, and navigates to a single target.

We can draw two conclusions from these findings. First, PRM-RL generalizes well to environments which are different from both our training environment and our previous testing environments. Second, PRM-RL is competitive with methods specifically designed for other environments.

\begin{figure*}[tb]
	\begin{center}
		\begin{tabular}{ccc}
            \subfloat[Effective vs Actual Radius]{\includegraphics[trim=0mm 0mm 0mm 0mm,clip,width=0.33\textwidth,height=4.5cm,keepaspectratio=true]{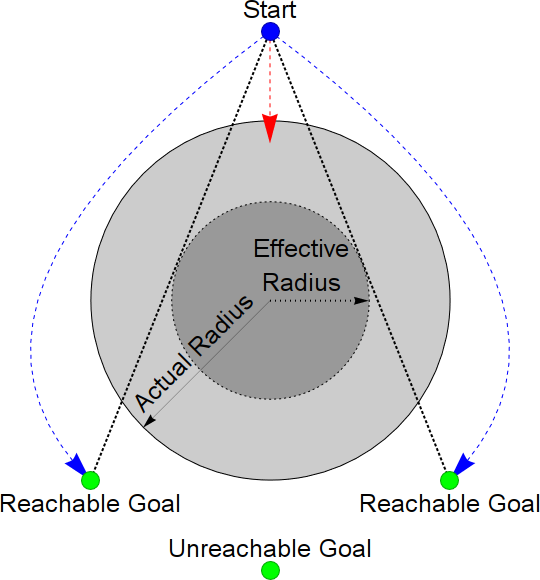}\label{fig:effective-radius}} 
			&
			\subfloat[Connections Possible with PRM-RL]{\includegraphics[trim=0mm 0mm 0mm
			0mm,clip,width=0.33\textwidth,height=4.5cm,keepaspectratio=true]{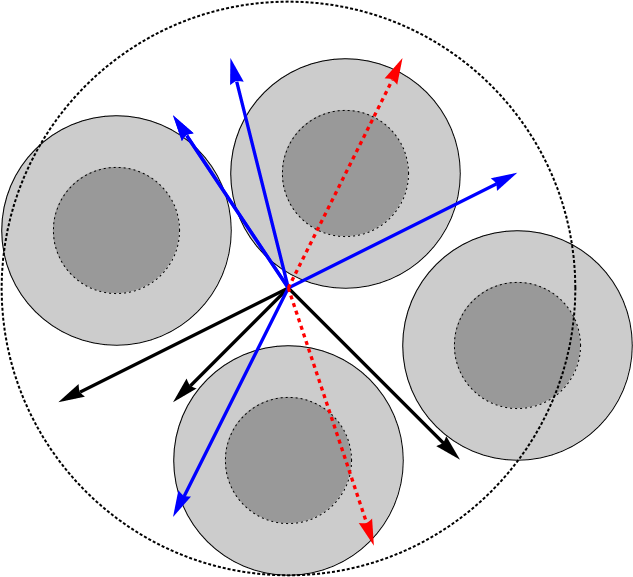}\label{fig:rl-addl-conn}} 
			&
			\subfloat[Increased Density of PRM-RL]{\includegraphics[trim=0mm 0mm 0mm 0mm,clip,width=0.33\textwidth,height=4.5cm,keepaspectratio=true]{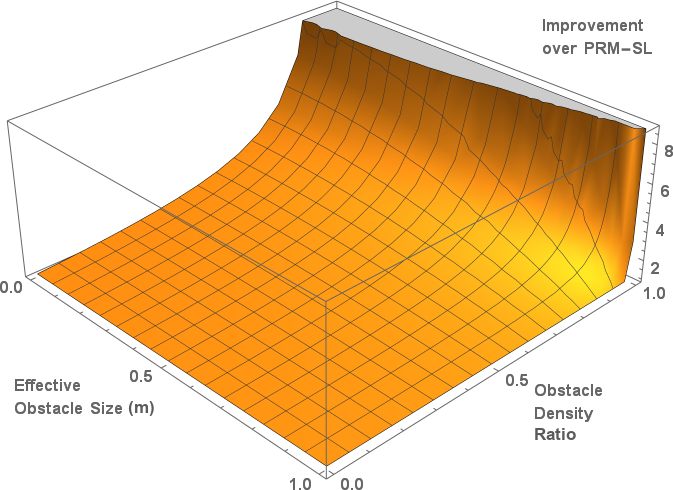}\label{fig:prm-rl-vs-sl}}
		\end{tabular}
		\caption{\small PRM-RL enables capturing valid samples not visible by line of sight. (a) The ability of RL to go around corners makes obstacles effectively smaller in configuration space. (b) This means more connections can be made for a given connectivity neighborhood. Solid black arrows represent valid connections for either PRM-SL or PRM-RL, dotted red arrows represent invalid connections for either method, and blue arrows indicate valid trajectories recovered by PRM-RL. (c) Compared to PRM-SL, PRM-RL recovers many more potential connections as obstacles grow denser and RL gets better. \label{fig:prm-rl-density}}
	\end{center}
\end{figure*}

\begin{figure*}[tb]
	\begin{center}
		\begin{tabular}{ccc}
            \subfloat[Effect of Noise in Corridors]{\includegraphics[trim=0mm 0mm 0mm 0mm,clip,width=0.3\textwidth,height=5cm,keepaspectratio=true]{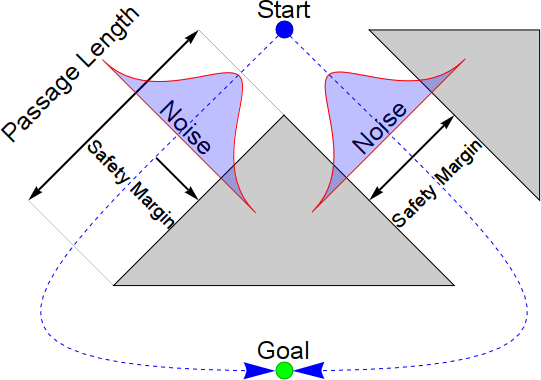}\label{fig:prm-rl-noise}} 
			&
			\subfloat[Reliability with 1 vs 2 Walls]{\includegraphics[trim=0mm 0mm 0mm
			0mm,clip,width=0.3\textwidth,height=5cm,keepaspectratio=true]{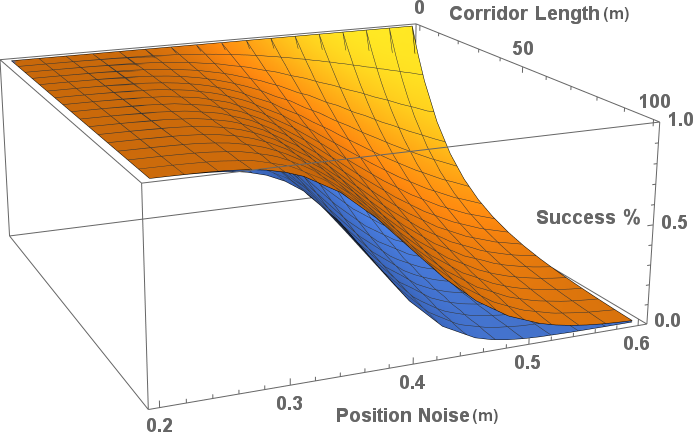}\label{fig:rl-reliable-vs-walls}} 
			&
			\subfloat[Overconfidence of PRM-SL vs PRM-RL]{\includegraphics[trim=0mm 0mm 0mm 0mm,clip,width=0.3\textwidth,height=5cm,keepaspectratio=true]{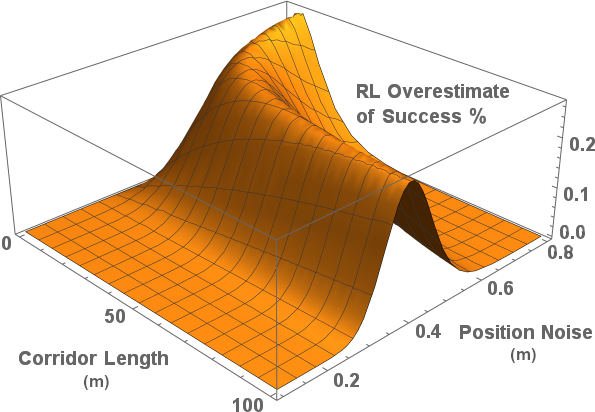}\label{fig:rl-overconfidence}}
		\end{tabular}
		\caption{\small PRM-RL enables capturing hazards in the environment difficult to learn with RL. (a) RL agents may learn to avoid obstacles, but not every location in the environment has identical clearance; on the left the robot can hug one wall, but on the right it must pass between two walls, so uncertainty in controls leads to two possible failure modes. (b) Therefore, as paths lengthen, the chance of navigating reliably drops faster in corridors than in wall hugging (blue region). (c) This leads to RL overconfidence in a regime between where both PRM-RL and RL are reliable and where they both are not; PRM-RL can encode this in the roadmap by deleting the right edge in (a). \label{fig:why-prm-rl-works}}
	\end{center}
\end{figure*}

\section{Analysis}
\label{sec:analysis}
In the previous section, we empirically established correlations between the local planner's competence and PRM-RL's resilience to noise. We also explored the contributions of sampling densities, success thresholds and obstacle map sources to the success of overall navigation. We concluded that 1) using an execution policy that is resilient to noise and avoids obstacles as a local planner improves the overall success of the hierarchical planner; 2) a success threshold of 100\% improves overall navigation success, 3) the upper bound to navigation success is not dependent on density but policy performance and robot type, and 4) using realistic obstacle maps, such as SLAM maps, as a basis for building roadmaps provides simulation results closer to reality.  

This section provides a deeper analysis of those empirical findings. Section \ref{sec:analysis-construction} analyzes the impact of local planner's obstacle avoidance and noise resilience on roadmap construction. Section \ref{sec:analysis-comp} examines the computational complexity of PRM-RL, Section \ref{sec:analysis-execution} discusses causes of failure for trajectories over multiple waypoints, and Section \ref{sec:future} discusses limitations of the method and future work.
\subsection{PRM-RL Roadmap Connectivity}
\label{sec:analysis-construction}
Unlike straight-line planners, RL agents can often go around corners and smaller obstacles; Fig.  \ref{fig:prm-rl-density} shows how this effectively transforms the configuration space to make obstacles smaller. While the agent never traverses these corner points, as they are not in $\cfree$, they nevertheless do not serve to block the agent's path, unlike central portions of a larger body which might block or trap the control agent in a local minimum. If we model this as an effective reduction in radius of a circular obstacle $f_{\pi}$ with respect to a policy $\pi$, and model the connection region as a disc filled with randomly sampled obstacles from $0\%$ to $100\%$ in total area density $\rho_{o}$, we can estimate an upper bound on connection success in the idealized case in which obstacles do not occlude and the chance of connection is just the complementary probability of encountering an obstacle over a region of space, $1 - \rho_{o}$. This corresponds to the looser bound $1-\rho_{o}{f_{\pi}}^2$ in the RL case. Therefore, a conservative estimate of the ratio of samples connected by PRM-RL to those connected by PRM-SL is:
\begin{equation}
\label{eq:effective-connectivity}
\frac{conn_{PRM-RL}}{conn_{PRM-SL}} = \frac{1-\rho_{o} {f_{\pi}}^2}{1-\rho_{o}}
\end{equation}
This simplified model indicates that it becomes harder to connect points as obstacle density increases, but PRM-RL has an increasing advantage over PRM-SL in connecting these difficult roadmaps as RL's ability to cut corners increases. Hence, in environments with dense obstacles, it makes sense to invest in execution policies that avoid obstacles really well, and use them as local planners. Alternately, it suggests that policies that can learn strategies for dealing with frequently occurring obstacle topologies, like box canyons and corners, are a fruitful area for future work. Because the branching factor of roadmap planning is proportional to the points in the connectivity neighborhood, planning cost can increase exponentially with connection radius, and following \cite{geraerts2004comparative}, we set neighborhood size empirically to balance the benefits of finding points within the effective navigation radius of the planner against the drawbacks of a connectivity neighborhood which contains so many points that planning becomes infeasible.

Conversely, PRM-RL does not connect nodes in a roadmap where a policy cannot navigate reliably. This is the key difference from PRM-SL, and is the cause for the upper limit on performance improvements as the roadmaps increase in density---the roadmaps are policy-bound, rather than sampling bound. One question to ask is why local control policies cannot learn to “drive safe” by repeatedly re-planning the path (e.g., via A* searches or variants). However, an analysis of how noise impacts the behavior of policies indicates that policies which do not memorize the environment may overestimate their ability to navigate because the hazards that they can observe locally may not represent an accurate picture of the hazards of the global environment. 

To see why, suppose a policy has learned to navigate safely in the presence of noise by maintaining a distance $d_{safety}$ from walls. Modeling time as discrete and assuming the robot is travelling at a constant speed, so that on each time slice the robot moves a constant distance $d_{step}$ units forward, let us further model sources of variability as transient Gaussian noise orthogonal to the robot's travel $\mathcal{N}_{pos}(0,\sigma _{pos})$ with zero mean and standard deviation $\sigma _{pos}$. This results in a probability of collision per step of $\frac{1}{2} \text{erfc}\left(\frac{d_{\text{safety}}}{\sqrt{2} \sigma _{\text{pos}}}\right)$ (the cumulative distribution function of the Gaussian noise model $\mathcal{N}_{pos}$ evaluated at $-d_{safety}$, expressed in terms of the complementary Gauss error function $\text{erfc}(x)$). Fig.  \ref{fig:prm-rl-noise} shows that when the robot is traveling in a narrow corridor it is twice as likely to collide as it does when hugging a wall, even though it may be maintaining $d_{safety}$ from any given wall at all times. Over a path of length $d_{corr}$, a conservative lower bound on the chance of collision rises exponentially with the number of steps it takes to traverse the path,
\begin{equation}
\label{eq:survival-probability}
P_{survival} = {\left(1-\frac{1}{2} \text{n}_{\text{walls}} \text{erfc}\left(\frac{d_{\text{safety}}}{\sqrt{2} \sigma  _{\text{pos}}} \right)\right)^{\frac{d_{\text{corr}}}{d_{\text{step}}}}}
\end{equation}
causing the narrow corridor case to become unsafe faster than the wall-hugging case as shown in the blue region of Fig.  \ref{fig:rl-reliable-vs-walls}.  This means that an RL policy that judges its safety based on locally observable features can overestimate the safety of a path in the region between where both PRM-RL and RL would succeed and where both PRM-RL and RL would fail (Fig.  \ref{fig:rl-overconfidence}). The same would be true of RL guided by PRM-SL based on clearance, such as in grid-based sampling \cite{unstuck-dinesh} and Safety PRM \cite{malone-iros-2013}. In this case, RL or RL guided by PRM-SL can make erroneous navigation choices, whereas PRM-RL simply does not add that link to the roadmap. While in theory an agent could be designed to cope with this specific issue, other scenarios can present similar problems: no local prediction of collision probability can give a true guarantee of trajectory safety. While this is an idealized analysis, our experience is that real-world environments can be more pathological---for example, at one site curtains behind a glass wall induced a persistent 0.5 meter error in the robot's perceived location, causing it to drive dangerously close to the wall for long distances despite the presence of a safety layer. PRM-RL provides an automated method to avoid adding those unreliable connections in the roadmap, resulting in the roadmaps that more sparsely connect but transfer more reliably to the robot.

\subsection{PRM-RL Computational Complexity}
\label{sec:analysis-comp}

In this section we assess the computational complexity of building a PRM-RL roadmap with known RL policy range $d_{\pi}$, sampling density $\rho_{\omega}$, workspace volume $V_\mathcal{W}$, and connection attempts $n_{\omega}$.
The cost of building a PRM with $n$ nodes is dominated by the cost of two distinct steps. The first step identifies potential edges by finding $m$ potential nearest neighbors for each of the $n$ nodes. While approaches such as PRM* can construct PRMs with $O(n \log n)$ edge tests, achieving this performance requires limits on $m$, such as connecting only the $m$ nearest neighbors or shrinking the radius of connection as $n$ increases \cite{karaman2011sampling}. Approaches using a fixed connection radius, such as Simplified PRM \cite{Kavraki98probabilisticroadmaps}, require $O(n^2)$ edge tests \cite{karaman2011sampling}. The second step evaluates the validity of each of the $n\,m$ candidate edges by performing collision checks along each edge, and adds valid edges to the graph.

PRM-RL samples nodes in the workspace $\mathcal{W}$ and attempts to connect all neighbors within range of the policy $d_{\myvec{\pi}}$, which can be determined empirically (see \cite{autorl}) and is independent of the workspace volume $V_{\mathcal{W}}$. Similarly, Section \ref{sec:results-connect} shows PRM-RL's performance shows diminishing returns beyond a  sampling density $\rho_{\omega}$, also independent of workspace volume. Capping sampling at density $\rho_{\omega}$ makes the number of nodes a function of the workspace volume, 
\begin{equation}
    n_{_\mathcal{W}} = V_{\mathcal{W}}\rho_{\omega} = O(d^{D_{\mathcal{W}}} \rho_{\omega}),
    \label{eq:npi}
\end{equation} 
where $D_\mathcal{W} \leq 3$ is the workspace dimensionality, and $d$ is the radius of the smallest $D_{\mathcal{W}}$-dimensional sphere that contains the workspace $\mathcal{W}.$
However, because the neighbor volume is a function of policy range and workspace dimensionality, 
$V_{\myvec{\pi}} \propto d_{\myvec{\pi}}^{D_{\mathcal{W}}},$
the number of neighbors 
\begin{equation}
\label{eq:mpi}
m_{\myvec{\pi}} = V_{\myvec{\pi}}\rho_{\omega} = O(d_{\myvec{\pi}}^{D_{\mathcal{W}}} \rho_{\omega})
\end{equation} is a variable independent from $n_{_\mathcal{W}},$ though their ratio is fixed for any given workspace. Let $d_{\myvec{\pi}} = cd$ for some constant $c \in \R{}$. More capable policies have $c$ closer to 1, while less capable policies have $c$ closer to 0. For simplicity, we assume that $0 < c \leq 1,$ when limiting the search radius, even if policy's radius might exceed the workspace's radius. Thus, \begin{equation}
    \frac{m_{\myvec{\pi}}}{n_{_\mathcal{W}}} \propto \frac{d_{\myvec{\pi}}^{D_{\mathcal{W}}}\rho_{\omega}}{d^{D_{\mathcal{W}}}\rho_{\omega}} = c^{D_{\mathcal{W}}} \leq 1.
    \label{eq:prop}
\end{equation}

Therefore we analyze PRM-RL with respect to nodes $n_{_\mathcal{W}}$ and neighbors $m_{\myvec{\pi}}$ in terms of the source variables that determine them: workspace volume $V_{\mathcal{W}}$, sampling density $\rho_{\omega}$, and effective range $d_{\myvec{\pi}}$. First, the edge identification step is $O(n_{_\mathcal{W}} \log n_{_\mathcal{W}})$ because the number of nearest neighbors given in Eq. \eqref{eq:mpi} is independent of the total number of nodes $n_{_\mathcal{W}}$ and can be found in $O(\log n_{_\mathcal{W}})$ with efficient approximate nearest neighbor searches, as described in \cite{karaman2011sampling}. From there,
\begin{align}
O(n_{_\mathcal{W}} \log n_{_\mathcal{W}}) &= O(d^{D_{\mathcal{W}}} \rho_{\omega} \log d^{D_{\mathcal{W}}}\rho_{\omega}),\; \text{due to \eqref{eq:npi}} \\
&= O(d^{D_{\mathcal{W}}} \rho_{\omega} \log d\rho_{\omega}), D_{_\mathcal{W}}\, \text{is constant.}
\label{eq:step1}
\end{align}
Second, for the edge validation phase, PRM-RL validates $m_{\myvec{\pi}} * n_{_\mathcal{W}}$ candidate edges with $n_{\omega}$ rollouts. To validate one edge rollout, PRM-RL performs on the order of $O(d_{\myvec{\pi}})$ collision checks. Therefore, the cost of adding neighbors to the roadmap is 
\begin{equation}
O(n_{_\mathcal{W}} m_{\myvec{\pi}} n_\omega d_{\myvec{\pi}}) = O(d^{D_{\mathcal{W}}} \rho_{\omega}^2 d_{\myvec{\pi}}^{D_{\mathcal{W}} + 1} n_\omega), 
\label{eq:step2}
\end{equation}
by substituting \eqref{eq:npi} and \eqref{eq:mpi} and rearranging. Combining Eq. \eqref{eq:step1} and \eqref{eq:step2}, we arrive at the total cost 
\begin{align}
    \,&O(d^{D_{\mathcal{W}}} \rho_{\omega} \log d\rho_{\omega} + d^{D_{\mathcal{W}}} \rho_{\omega}^2 d_{\myvec{\pi}}^{D_{\mathcal{W}} + 1} n_\omega)  = \\
    \,&O(d^{D_{\mathcal{W}}} d_{\myvec{\pi}}^{D_{\mathcal{W}} + 1} \rho_{\omega}^2 n_\omega) \label{eq:com}
\end{align}
because for a typical map $\rho_{\omega} d_{\myvec{\pi}}^{D_{\mathcal{W}}}$ dominates $\log d\rho_{\omega}.$ 

Eq. \eqref{eq:com} exposes the following power sources:
\begin{itemize}
    \item Complexity is $O({d_{\pi}}^{2D_{\mathcal{W}} + 1})$ in the policy range, so local planners that can reliably cover longer distances increase the computational cost of the roadmap. We recommended choosing a shorter connection distance $d_{\omega} < d_{\pi}$ even if the policy is capable of longer connections. 
    \item When the workspace is much larger than the reach of the policy ($0 < c \ll 1$), the complexity is almost linear in workspace volume.  
    \item Complexity is linear in connection attempts $n_\omega$.
    \item Complexity is quadratic in sampling density $\rho_{\omega}$, making it worthwhile to assess the limiting sampling density before building large numbers of roadmaps.
\end{itemize}

Each edge connection attempt is independent, so roadmap building can be parallelized up to the expected number of samples $V_\mathcal{W} \rho_{\omega}$. If parallel rollouts are performed instead of early termination, this can be parallelized further by an additional factor of $n_\omega$. So, given $n_p \leq V_\mathcal{W} \rho_{\omega} n_\omega$ processors, the effective time complexity can be reduced up to $O(\frac{V_\mathcal{W} \rho_{\omega} n_{\omega}}{n_p} \rho_{\omega} {d_{\pi}}^{D_\mathcal{W} + 1} ),$ possibly alleviating some of the time cost of increased sampling.

Last, note that when using early termination, increasing the success threshold $p_s$ often (but not always) reduces the required number of connection attempts $n_s$. In the worst case scenario where we require $p_s = 0.5$, early termination can at best cut $n_s$ to $\frac{n_{\omega}}{2}$, but as $p_s$ increases the number of failures needed to exclude an edge, $\frac{n_{\omega}}{1-p_s}$, drops toward 1. Conversely, if navigation is successful then the full $n_{\omega}$ samples need to be collected for an edge; the distribution of successes and failures thus has a large effect on the cost. One way to control this cost to reduce the max connection distance $d_{\omega}$ to less than the effective policy navigation distance $d_{\pi}$; in this case the agent is more often expected to succeed, and $n_{\omega}$ can potentially be reduced. We have observed that these tradeoffs can significantly  affect the cost of a run, but must be studied empirically for the environments of interest.

\subsection{PRM-RL Trajectory Execution}
\label{sec:analysis-execution}
Because PRM-RL construction calculates the probability of success before adding an edge, we can estimate the expected probability of success of a long-range path that passes through several waypoints. Recall that to add an edge to the roadmap we collect $n_{\omega} = 20$ Monte Carlo rollouts and require an observed proportion of successes $p_s$ typically of $90\%$ and $100\%$. Given that expected probability of success of a Bernoulli trial observing $n_s$ successes out of $n_{\omega}$ samples is 
\begin{equation}
\label{eq:success-estimator}
\mathbb{E}[p_{s}] = \frac{n_{s} + 1}{n_{\omega} + 2}
\end{equation} 
\cite{DBLP:journals/corr/abs-1105-1486}, the actual probability of successful navigation $p_n$ over an edge with $p_s = 100\%$ successful samples is $95.5\%$, and similarly $p_n = 86.3\%$ for thresholds of $p_s = 90\%$. Extrapolating over the sequence of edges in a PRM-RL path, the expected success rate is ${p_{n}}^{n_w}$ where $n_w$ is number of waypoints. In \cite{prm-rl} we observe PRM-RL paths with  $10.25$ waypoints averaged over our three deployment maps, yielding an estimated probability of success of $22.0\%$ for the $90\%$ threshold and $62.3\%$ for the $100\%$ threshold. Therefore, for the lengths of paths we observe in our typical deployment environments, the $100\%$ threshold improves PRM-RL's theoretical performance to the point that it is more likely to succeed than not, which is what we observe empirically.

\subsection{Limitations}
\label{sec:future}
While AutoRL can handle moving obstacles \cite{autorl}, PRM roadmaps remain static after construction, causing two failure modes. First, PRM-RL does not re-plan. If dynamic obstacle avoidance steers a local planner closer to a subsequent waypoint, PRM-RL could re-plan and provide that waypoint as the next waypoint. Second, large changes in the environment can invalidate edges or create new paths (e.g., adding/removing a wall). Re-planning with a roadmap update could handle this scenario (for example, with iterative reshaping \cite{iterative-3d}).

Another limitation of PRM-RL is that it requires a map. With a sufficiently good local policy, a SLAM algorithm, and an incrementally updatable PRM, it would be possible to build a PRM online by progressively exploring an environment and building the roadmap and SLAM map together. However, while adding online features to a roadmap are certainly feasible, developing an exploration policy is challenging in its own right, and goes hand in hand with improving the quality of the local planner so it can be trusted to execute reliably. 

This work focuses on evaluating roadmap construction and performance, so we leave replanning, exploration policies, and online map building for future work.

\section{Conclusion}
\label{sec:conclusion}
We presented PRM-RL, a hierarchical planning method for long-range navigation that combines sampling-based path planning with RL agent as a local planner in very large environments. The core of the method is that roadmap nodes are connected only when the RL agent can connect them consistently in the face of noise and obstacles. This extension of \cite{prm-rl} contributed roadmap construction and robot deployment algorithms, along with roadmap connectivity, computational complexity, and navigation performance analysis. We evaluated the method on a differential drive and a car model with inertia used on floormaps from five building, two noisy obstacle maps, and on three physical testbed environments. 

We showed that 1) the navigation quality and resilience to noise of the execution policy directly transfers to the hierarchical planner; 2) a 100\% success threshold in roadmap construction yields both the highest quality and most computationally efficient roadmaps; 3) building roadmaps from the noisy SLAM maps that the robot uses at execution time virtually closes the sim2real gap, yielding simulation success rates very similar to those observed on robots. PRM-RL with SLAM maps embed information into the roadmap that the robot uses at execution time, providing a better estimate of performance on the robot. Failure modes include pathologies of the local policy, poorly positioned samples, and sparsely connected roadmaps. In future work, we will examine improved policies with more resistance to noise, better sampling techniques to position the samples strategically, and techniques for improving map coverage with better localization and obstacle maps. 

\section*{Acknowledgments}
The authors thank Jeffrey Bingham, James Davidson, Brian Ichter, Ken Oslund, Peter Pastor, Oscar Ramirez, Chad Richards, Lydia Tapia, Alex Toshev, and Vincent Vanhoucke for helpful discussions and contributions to this project. We also thank the editors and reviewers for their detailed reviews and thoughtful comments. 

\appendix[Table of Symbols]
\label{app:symbols}
\begin{table}[H]
\scriptsize
\label{tab:constants}
    \centering
    \begin{tabular}{l|c|l}
    \hline \hline
    Symbol                          & Units or Domain                   & Meaning \\ \hline
    $\cspace$     & $\mathbb{R}^{D_c}$                & Configuration space of dimension $D_c$\\ 
    $\cfree$       & $\mathbb{R}^{D_c}$                & Free portion of configuration space \\ 
    $S$             & $\mathbb{R}^{{D_c}+{D_t}}$        & State space of robot plus task state\\ 
    $T$             & $\mathbb{R}^2 \times \mathbb{S}^1$   & Task space for navigation\\ 
    $\tfree$                      & $\mathbb{R}^{D_\mathcal{W}}$                & Free portion of the task space\\ 
    $\mathcal{W}$    & $\mathbb{R}^{D_\mathcal{W}}$   & Physical workspace of dimension $D_\mathcal{W}$ \\ 
    $O$             & $\mathbb{R}^{D_o}$    & Observation space of dimension $D_o$\\ 
    $A$             & $\mathbb{R}^{D_a}$                & Action space of dimension $D_a$\\ 
    $D$          & $S \times A$          & Task dynamics $\rightarrow \cspace$\\ 
    $N$          & $\cspace \times A$   & Noise model $\rightarrow O$ or $A$\\ 
    $R$          & $O$  & Reward model $(G, r)$ \\
    $G$         & $\cspace$           & True objective $\rightarrow \mathbb{R}$ \\ 
    $r$         & $O$ parametrized with $\myvec{\theta}$                 & Dense reward $\rightarrow \mathbb{R}$ \\ 
    $r_\text{name}$  & $O$ parametrized with $\myvec{\theta}_\text{name}$          & Named reward component $\rightarrow \mathbb{R}$ \\
    $\gamma$     & $[0..1]$                          & Discount \\
    $\mathbb{I}$        & $\mathbb{B}$      & Indicator function $\rightarrow \{1, 0\}$\\ 
    $L(\myvec{x})$      & $\cspace$   & Task predicate \\ 
    $F_{s}(\myvec{x})$  & $\cspace$   & Sensor w/ dyn. $D_{s}(\myvec{x})$ 
                                                                            \& noise $\mathcal{N}_{s}$\\
    $F_{a}(\myvec{x}, \myvec{a})$   & $\cspace \times A$              & Action w/ dyn. $D_{a}(\myvec{x})$ 
                                                                            \& noise $\mathcal{N}_{a}$\\
    $\myvec{p}_i$                   & $\cfree$                        & Waypoint $i$ on path $\mathcal{P}$ \\
    $\myvec{x_{i}}$                 & $\cspace$                       & Point $i$ along trajectory $\mathcal{T}$ \\
    $\myvec{x_{S}}$                 & $\cfree$                        & Start state \\
    $\myvec{x_{G}}$                 & $\cfree$                        & Goal state \\
    $d_{G}$                         & meters                            & Goal success distance \\
    $\mathcal{K}_{\omega}$          & $\mathbb{Z}+$                     & Max trajectory execution steps \\
    $\ac$                           & $\mathbb{R}^2 \times \mathbb{S}^1$
                                                                        & Diff drive action $(v, \phi)$ lin, ang vel. \\      
    $\overline{V_i V_j}$            & $\cspace$                       & Line from $V_i$ to $V_j$ in graph $(V, E)$ \\
    $n_{\omega}$                    & $\mathbb{Z}+$                     & Num. edge connection attempts \\
    $n_{s}$                         & $\mathbb{Z}*$                     & Num. observed connection successes \\
    $p_{s}$                         & $[0..1]$                          & Edge connection success threshold \\
    $d_{\omega}$                    & meters                            & Max attempted edge connection dist. \\
    $d_{\pi}$                       & meters                            & Policy $\pi$ effective nav. distance \\
    $f_{\pi}$                       & $[0..1]$                          & Policy $\pi$ effective obst. shrinkage \\
    $\rho_{\omega}$                 & $\text{points}/\text{meters}^2$   & Sampling density per meter \\
    $n$                             & points                            & Number of points to sample \\
    $p_{n}$                         & $[0..1]$                          & Probability of successful navigation \\
    $n_w$                           & $\mathbb{Z}+$                     & Number of waypoints on a path \\
    $V_\mathcal{W}$                           & $\text{meters}^2$                 & Volume of the workspace \\
    $D_\mathcal{W}$                           & $\mathbb{Z}+$                     & Workspace dimensionality \\
    $n_{_\mathcal{W}}$                           & $\mathbb{Z}+$                     & Nodes in the workspace \\
    $m_\pi$                           & $\mathbb{Z}+$                     & Neighbors of a node \\
    $n_p$                           & $\mathbb{Z}+$                     & Number of processors \\
    $\sigma_{x}$                    & $\mathbb{R}+$                     & $\gauss{0}{\sigma_x}$ noise for $x = l,g,v,a$ \\
    $\theta_n$               & $\mathbb{Z}+$          & Observation trace length \\
    \hline \hline
    \end{tabular}
\end{table}

\appendix[Training Hyperparameters]
\label{app:training}
Both actor and critic use the AdamOptimizer with $\beta_1 = 0.9$,  $\beta_2 = 0.999$, $\epsilon = 1e-08$; the actor's learning rate is $1e-05$ and the critic's is $0.0005$. The actor uses DQDA gradient clipping and the critic uses $\gamma = 0.995$ with the Huber loss for temporal difference errors. $10,000$ initial stabilization steps are followed by a soft target network update of $0.0001$ on every step. Our training batch size is $512$ and our replay buffer has a capacity of $0.5$ million. We train for $5$ million steps, but save policies every $25,000$ steps and select the best policy over the run.

\ifCLASSOPTIONcaptionsoff
  \newpage
\fi



%
%
%

\small
\bibliographystyle{abbrv}
\bibliography{literature}  

%

\begin{IEEEbiography}[{\includegraphics[width=1in,height=1.25in,clip,keepaspectratio]{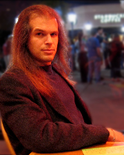}}]{Anthony Francis}
is a Senior Software Engineer at Robotics at Google specializing in reinforcement learning for robot navigation. Previously, he worked on emotional long-term memory for robot pets at Georgia Tech's PEPE robot pet project, on models of human memory for information retrieval at Enkia Corporation, and on large-scale metadata search and 3D object visualization at Google.
He earned his B.S. (1991), M.S. (1996) and Ph.D. (2000) in Computer Science from Georgia Tech, along with a Certificate in Cognitive Science (1999). He and his colleagues won the ICRA 2018 Best Paper Award for Service Robotics. Anthony's work has been featured in the New York Times.
\end{IEEEbiography}

\excise{\begin{IEEEbiography}[{\includegraphics[width=1in,height=1.25in,clip,keepaspectratio]{figures/headshot-francis.png}}]{Anthony Francis}
is a Senior Software Engineer at Robotics at Google specializing in reinforcement learning for robot navigation. Previously, he worked on emotional long-term memory for robot pets at Georgia Tech's PEPE robot pet project, on models of human memory for information retrieval at Enkia Corporation, and on large-scale metadata search and 3D object visualization at Google.
He earned his B.S. (1991), M.S. (1996) and Ph.D. (2000) in Computer Science from Georgia Tech, along with a Certificate in Cognitive Science (1999). He has been a member of IEEE since 2014. He and his colleagues won the ICRA 2018 Best Paper Award for Service Robotics for their paper on PRM-RL. He is an author on a dozen peer-reviewed publications and an inventor on a half-dozen patents. He's published a dozen short stories and four novels, including the EPIC eBook Award-winning \textit{Frost Moon}; his popular writing on robotics includes articles in \textit{Star Trek Psychology}, in \textit{Westworld Psychology}, and the Google AI blogpost ``Maybe your computer just needs a hug.'' He lives in San Jose with his wife and cats, but his heart will always belong in Atlanta.
\end{IEEEbiography}
}

\begin{IEEEbiography}[{\includegraphics[width=1in,height=1.25in,clip,keepaspectratio]{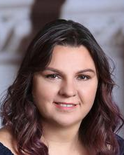}}]{Aleksandra Faust} is a Staff  Research Scientist at Robotics at Google, specializing in robot motion planning and reinforcement learning. 
She earned a Ph.D. in Computer Science at the University of New Mexico (with distinction), and a Master's in Computer Science from the University of Illinois at Urbana-Champaign.
Her work has been featured in the New York Times, ZdNet, and was awarded Best Paper in Service Robotics at ICRA 2018.
\end{IEEEbiography}

\begin{IEEEbiography}[{\includegraphics[width=1in,height=1.25in,clip,keepaspectratio]{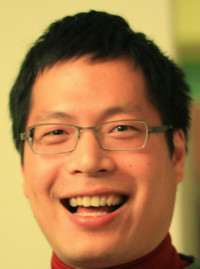}}]{Hao-Tien (Lewis) Chiang}
received a M.S. in Physics from the University of Mexico and worked on quantum algorithms. He is currently a Ph.D candidate in the Computer Science Department of the University of New Mexico. He was an organizer of the Third Machine Learning in Planning and Control Workshop at ICRA 2018, the Becoming a Robot Guru 2 workshop at RSS 2016, and the Becoming a Robot Guru 3 workshop at WAFR 2018. He is currently a PhD Intern at Robotics at Google.
\end{IEEEbiography}

\begin{IEEEbiography}[{\includegraphics[width=1in,height=1.25in,clip,keepaspectratio]{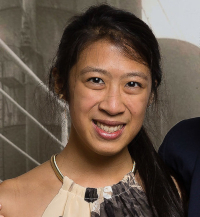}}]{Jasmine Hsu}
received a B.A. in Cognitive Science from the University of Virginia and a M.S. in Computer Science from New York University.  Jasmine Hsu previously worked in the defense industry and has been at Robotics at Google since 2016.  Her previous work has been focused on reinforcement learning for grasping, learning representations, and currently motion-planning.
\end{IEEEbiography}

\begin{IEEEbiography}[{\includegraphics[width=1in,height=1.25in,clip,keepaspectratio]{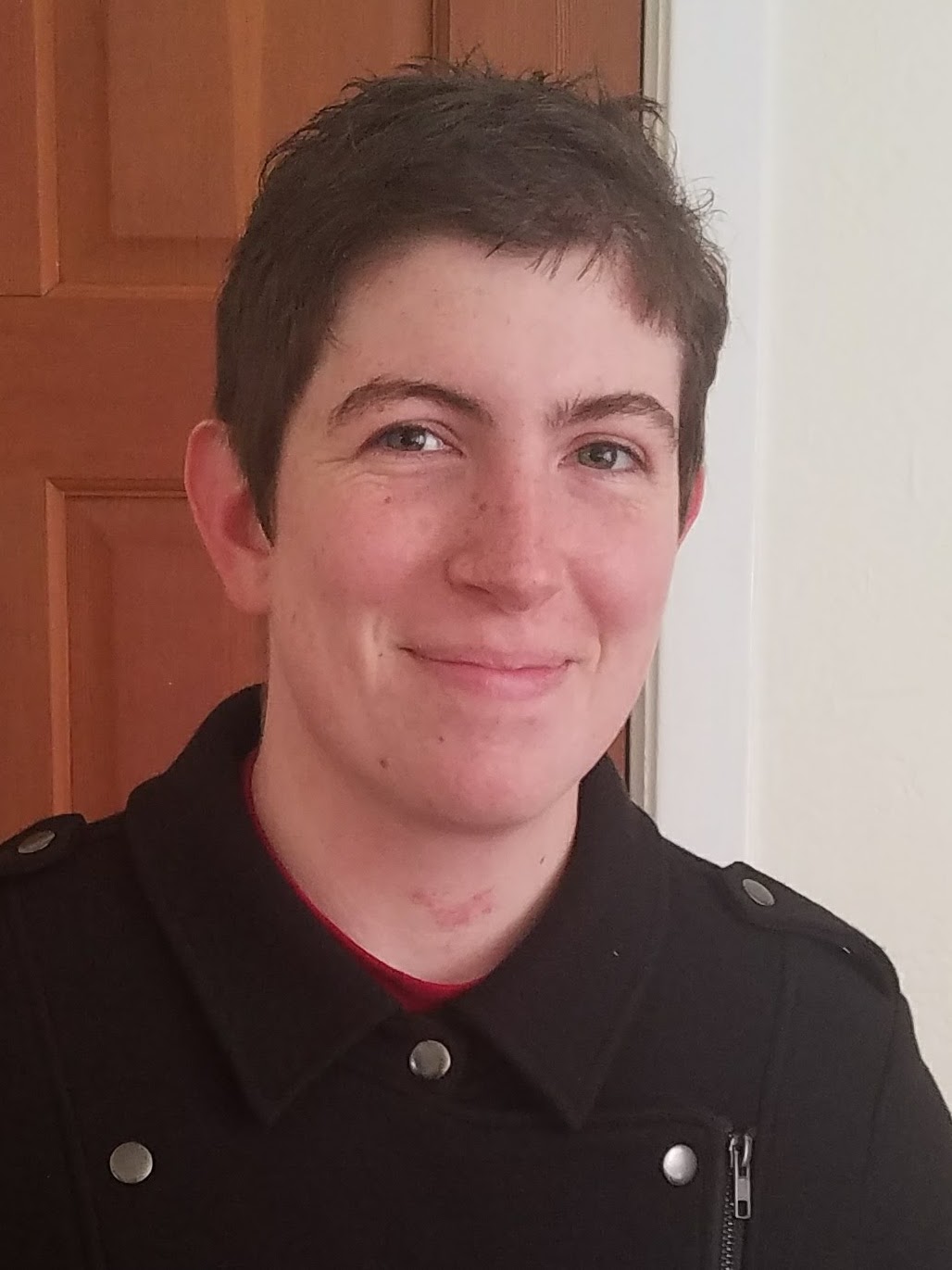}}]{J. Chase Kew}
received a B.S. in Computer Science and Mechanical Engineering from the California Institute of Technology. She is a Software Engineer at Robotics at Google working on machine learning for robotic navigation.
\end{IEEEbiography}


\begin{IEEEbiography}[{\includegraphics[width=1in,height=1.25in,clip,keepaspectratio]{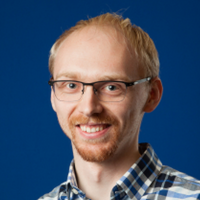}}]{Marek Fiser}
is a Software Engineer in the Robotics at Google team working on systems that can learn navigation policies with reinforcement learning, including creation and integration of simulated environments, designing and training agents using RL, and deploying learnt policies on real robots.
Marek has a Master's in Computer Graphics from Purdue University. His work was awarded Best Paper in Service Robotics in 2018.
\end{IEEEbiography}

\begin{IEEEbiography}[{\includegraphics[width=1in,height=1.25in,clip,keepaspectratio]{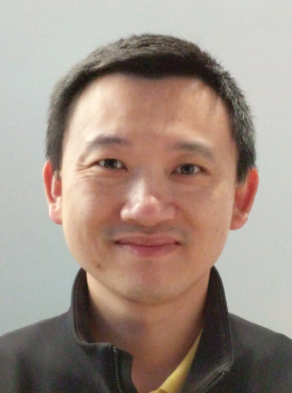}}]{Tsang-Wei Edward Lee}
received a B.S in Computer Science and Electrical Engineering from the University of California, Riverside.  He is a Test Engineer for Genesis10 working at Robotics at Google designing test plans, supporting robot operations and conducting on robot experiments.
\end{IEEEbiography}




\end{document}